\documentclass{article} 
\usepackage{iclr2025_conference,times} 

\usepackage{microtype}
\usepackage{graphicx}
\usepackage{subcaption}
\usepackage{booktabs} 
\usepackage{xcolor}
\usepackage{amsmath}
\usepackage{amssymb}
\usepackage{mathtools}
\usepackage{amsthm}
\usepackage{tcolorbox}
\usepackage{listings}
\usepackage{xcolor}

\definecolor{codegreen}{rgb}{0,0.6,0}
\definecolor{codegray}{rgb}{0.5,0.5,0.5}
\definecolor{codepurple}{rgb}{0.58,0,0.82}
\definecolor{backcolour}{rgb}{0.95,0.95,0.92}

\lstdefinestyle{mystyle}{
    backgroundcolor=\color{backcolour},   
    commentstyle=\color{codegreen},
    keywordstyle=\color{magenta},
    numberstyle=\tiny\color{codegray},
    stringstyle=\color{codepurple},
    basicstyle=\ttfamily\footnotesize,
    breakatwhitespace=false,         
    breaklines=true,                 
    captionpos=b,                    
    keepspaces=true,                 
    numbers=left,                    
    numbersep=5pt,                  
    showspaces=false,                
    showstringspaces=false,
    showtabs=false,                  
    tabsize=2
}

\usepackage{hyperref}

\usepackage{pgfplots}
\pgfplotsset{compat = newest}
\usepackage{multirow}

\usepackage{amsmath}
\usepackage{amssymb}
\usepackage{mathrsfs}
\usepackage{mathtools}
\usepackage{amsthm}
\usepackage{algorithm}
\usepackage{algpseudocode}
\usepackage{listings}
\usepackage{makecell}
\usepackage{colortbl}
\usepackage{color}
\usepackage{wrapfig}
\usepackage{cancel}
\usepackage{soul,xcolor}
\usepackage{pifont}
\usepackage{tikz}
\usetikzlibrary{shapes, positioning, arrows.meta, calc, decorations.pathmorphing, quotes}

\usepackage[capitalize,noabbrev]{cleveref}

\theoremstyle{plain}

\theoremstyle{definition}

\theoremstyle{remark}

\usepackage[textsize=tiny]{todonotes}

\newcommand{\alink}[1]{\href{#1}{paper-link}}

\usepackage{enumitem}
\usepackage{ amssymb }

\definecolor{codebg}{rgb}{0.95,0.95,0.95}
\definecolor{codeblue}{rgb}{0.13,0.13,1}
\definecolor{codegreen}{rgb}{0,0.5,0}
\definecolor{codegray}{rgb}{0.5,0.5,0.5}
\definecolor{codered}{rgb}{0.7,0,0}

\definecolor{citecolor}{HTML}{0071BC}
\definecolor{linkcolor}{HTML}{D32F2F}
\definecolor{cellcolor}{HTML}{E3F2FD}
\definecolor{red}{HTML}{D32F2F}
\definecolor{magenta}{HTML}{D81B60}
\usepackage{url}
\hypersetup{colorlinks=true, linkcolor=linkcolor, citecolor=citecolor,urlcolor=magenta}

\usepackage{amsmath,amsfonts,bm}

\def\eqref#1{equation~\ref{#1}}

\def\1{\bm{1}}

\DeclareMathAlphabet{\mathsfit}{\encodingdefault}{\sfdefault}{m}{sl}
\SetMathAlphabet{\mathsfit}{bold}{\encodingdefault}{\sfdefault}{bx}{n}

\renewcommand{\cite}[1]{\citep{#1}}
\usepackage{url}
\usepackage{xcolor}

\title{Diabetica: Adapting Large Language Model to \\ Enhance Multiple Medical Tasks in\\ Diabetes Care and Management}

\author{%
    \bf Lai Wei$^{1,*}$, Zhen Ying$^{1,*}$, Muyang He$^{1,}$\thanks{Equal contribution.}\hspace*{0.4em}, Yutong Chen$^{1}$, Qian Yang$^{3}$, Yanzhe Hong$^{1}$  \\[0.3em]
    \bf Jiaping Lu$^{4}$, Kaipeng Zheng$^{1,2}$, Shaoting Zhang$^{1}$, Xiaoying Li$^{1}$\\[0.3em]
    \bf Weiran Huang$^{1,2,\dag}$,
    Ying Chen$^{1,}$\thanks{Correspondence to Weiran Huang and Ying Chen.} 
    \\[0.6em]
    $^1$ Qing Yuan Research Institute, School of Computer Science, Shanghai Jiao Tong University; \\
    \hspace*{0.7em}Ministry of Education Key Laboratory of Metabolism and Molecular Medicine, 
    Department of\\ 
    \hspace*{0.7em}Endocrinology and Metabolism, Zhongshan Hospital, Fudan University \\[0.3em]
    $^2$ Shanghai Innovation Institute \\[0.3em]
    $^3$ Department of Endocrinology, Fifth People’s Hospital of Shanghai, Fudan University \\[0.3em]
    $^4$ Department of Endocrinology and Metabolism, Qingpu Branch of Zhongshan Hospital, \\
    \hspace*{0.7em}Affiliated to Fudan University 
}

\iclrfinalcopy 
\begin{document}

\maketitle
\renewcommand{\thefootnote}{\fnsymbol{footnote}}
\setcounter{footnote}{2}


\begin{abstract}
Diabetes is a chronic disease with a significant global health burden, requiring multi-stakeholder collaboration for optimal management. Large language models (LLMs) have shown promise in various healthcare scenarios, but their effectiveness across diverse diabetes tasks remains unproven. Our study introduced a framework to train and validate diabetes-specific LLMs. We first developed a comprehensive data processing pipeline that includes data collection, filtering, augmentation and refinement. This created a high-quality, diabetes-specific dataset and evaluation benchmarks from scratch. Fine-tuned on the collected training dataset, our diabetes-specific LLM family demonstrated state-of-the-art proficiency in processing various diabetes tasks compared to other LLMs. Furthermore, clinical studies revealed the potential applications of our models in diabetes care, including providing personalized healthcare, assisting medical education, and streamlining clinical tasks. 
Generally, our introduced framework helps develop diabetes-specific LLMs and highlights their potential to enhance clinical practice and provide personalized, data-driven support for diabetes management across different end users.
Our codes, benchmarks and models are available at
{\url{https://github.com/waltonfuture/Diabetica}}.
\end{abstract}

\section{Introduction}

Diabetes mellitus, affecting 10\% of the global population, stands as one of the most prevalent chronic diseases worldwide. Despite global efforts, challenges such as a shortage of diabetes specialists, uneven distribution of medical resources, low diabetes knowledge awareness, and inadequate self-management capabilities persist. These issues lead to poor glycemic control, resulting in substantial mortality and social burden~\cite{guan2023artificial}. With diabetes prevalence projected to rise to 643 million by 2030 and 783 million by 2045~\cite{sun2022idf}, current diabetes care systems would not be able to scale to meet the increasing demand. Optimizing diabetes management requires multi-stakeholder collaboration to strengthen specialist training and improve patient self-management capabilities. Therefore, there is an urgent need for a novel diabetes management instrument with accessibility, reliability and efficiency.

Recent developments in large language models (LLMs) have shown a rapid progress, equipped with advanced language comprehension capabilities and the ability to handle complex linguistic tasks~\cite{openai2023gpt4, anil2023palm, team2023gemini, bai2023qwen, hurst2024gpt}. 
Recent research shows that general models fine-tuned with medical datasets can yield performance on par with commercial models with larger scales, offering a viable method for delivering cost-effective and transparent clinical support~\cite{zhang2024closing, van2024adapted}. Additionally, the medical field can be further divided into departments with unique disease spectrums, general medical LLMs trained on broad medical data may fail to capture in-depth domain-specific knowledge so that perform inadequately when confronted with specialized clinical questions. While several open-source model architectures were proposed for specialized medical domain~\cite{chen2024ffa,zhou2024pre}, models specifically addressing diabetes are rarely reported~\cite{li2024integrated}, primarily due to the lack of high-quality datasets and appropriate paradigms. Thus, it is crucial to develop a tailored LLM for diabetes, which holds remarkable promise in advancing personalized, data-driven support for both patients and healthcare professionals.

Moreover, due to the life-critical nature of healthcare applications, using medical LLMs necessitates objective and comprehensive evaluation of the models’ performance and capabilities. 
However, there is a lack of benchmarks for diabetes specialties. Clinical practice also differs significantly from simply answering examination questions correctly, and finding appropriate benchmarks to gauge the clinical potential of LLMs is a substantial challenge~\cite{thirunavukarasu2023large}. Therefore, to validate the effectiveness and utility of specific models, there is an urgent need to provide a comprehensive diabetes assessment framework that balances both laboratory and clinical practice performance. 

\begin{figure}[t]
\centering
\includegraphics[width=1\textwidth]{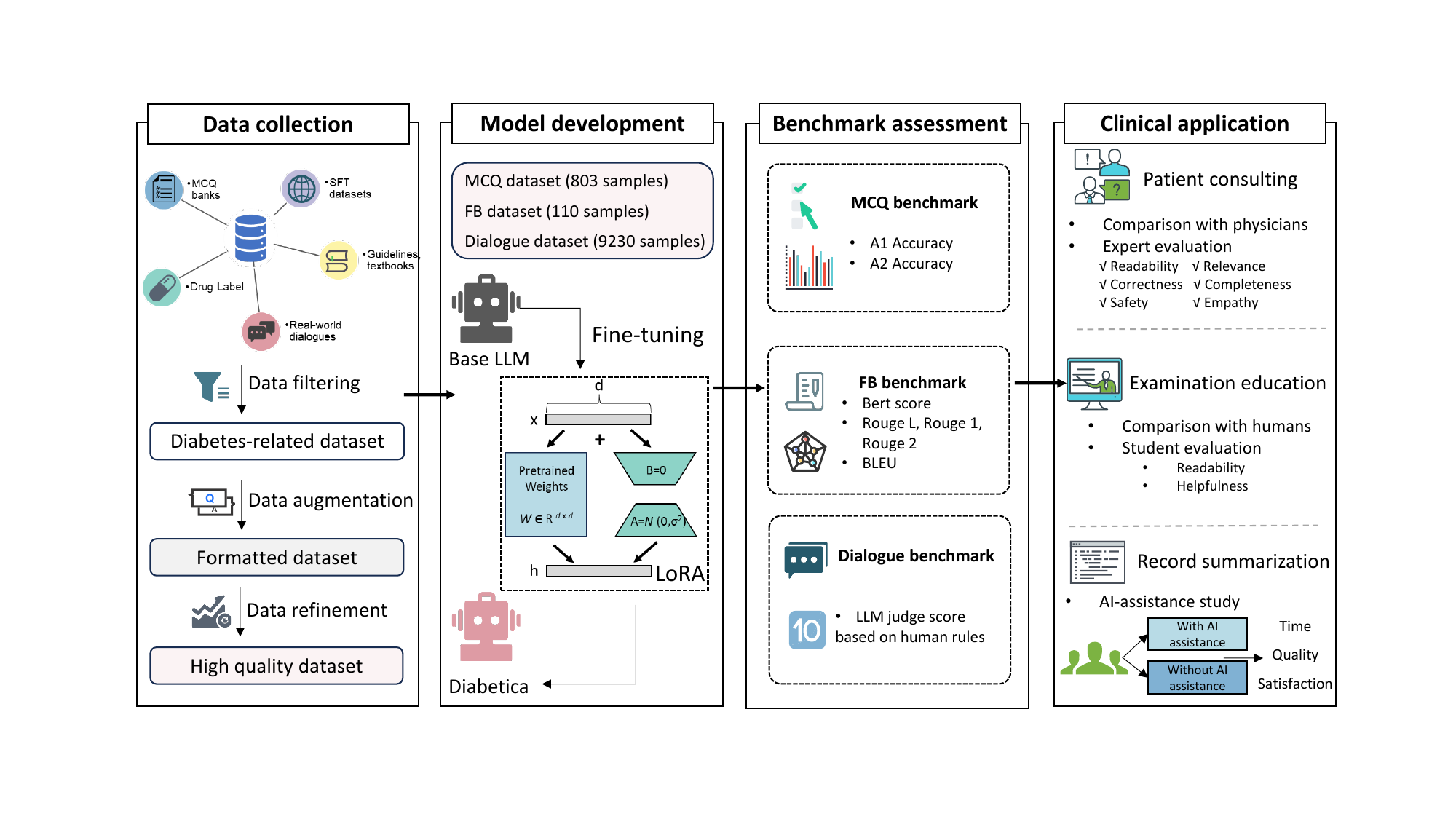}
\vspace{-0.9cm}
\caption{Overall study design. (a) Training data was collected from various sources. Data processing was then conducted to get the final diabetes-related, formatted, and high-quality dataset. (b) Fine-tuning was applied for developing Diabetica. (c) We compared the performance of different LLMs on MCQ benchmark, FB benchmark, and dialogue benchmark. (d) Our model was then evaluated in three clinical applications: medical consulting, examination education, and clinical record summarization.}
\label{fig:1}
\end{figure}

To this end, we introduced a reproducible paradigm shown in Figure~\ref{fig:1} to develop a specialized LLM called \emph{Diabetica} that could handle a wide range of diabetes-related tasks.
In particular, we curated a diverse dataset from multiple sources and employed different data processing approaches to guarantee the high quality.
Then, Diabetica was trained using this dataset with a self-distillation method to enhance alignment with human preferences and diabetes-specialized knowledge.
In our experiments, Diabetica demonstrated superior performance on various diabetes-related assessment benchmarks, including multiple-choice questions, fill-in-the-blank questions, and dialogue tasks, surpassing other open-source models of similar size and even matching or exceeding state-of-art proprietary LLMs. Furthermore, clinical evaluations have confirmed the effectiveness of our model in patient consulting, medical education, and optimizing clinical workflows, showcasing its potential for diverse applications in diabetes management facing different end users.

To summarize, our study makes several key contributions as follows:
\begin{itemize}
[itemsep=0pt,topsep=0pt,leftmargin=0.5cm]
\item We present a reproducible paradigm for developing specialized medical LLMs. Our approach demonstrated how to effectively leverage open-source models, curate high-quality disease-specific datasets, and fine-tune models in a particular medical domain. 
\item We have designed and created comprehensive evaluation benchmarks specifically for the diabetes field, which encompass a wide range of tasks with diverse formats. The assessment results demonstrated the state-of-the-art performance of our models in comprehending and executing diabetes-related tasks. 
\item We conducted a series of clinical studies to evaluate the LLM’s efficacy in real-world settings, which showed the potential practical applications of our model in diabetes care and how they could contribute to providing personalized healthcare, assisting medical education, and streamlining clinical tasks. 
\end{itemize}

\section{Diabetica}


In this section, we aim to develop a specialized LLM for diabetes-related applications, i.e., \emph{Diabetica}, by employing various and systematical data processing strategies as follows.


\subsection{Data collection}

We collect a wide range of diabetes-specific language data from various sources, including public multi-choice questions and medical SFT datasets, as well as our private in-house dataset derived from guidelines, textbooks, drug labels and real-world dialogues.  
Please refer to Appendix~\ref{app:dataset} for details. 

\subsection{Data filtering}

We first conducted data filtering, including keywords filtering and deduplication, to construct a diabetes-related dataset. 

\textbf{Keywords filtering.} To extract diabetes-related questions from our endocrinology MCQ dataset, we developed a keyword filtering system that incorporated both positive and negative matching. For positive matching, we identified and used keywords such as ‘diabetes’, ‘DKA’ (diabetic ketoacidosis), ‘blood sugar’, ‘HbA1c’ (hemoglobin A1c), ‘pancreas’, as well as the names of commonly prescribed diabetes medications. For negative matching, we crafted a specific list of exclusion keywords after thoroughly reviewing the dataset content. These exclusion keywords included terms like ‘insulinoma’, ‘short bowel syndrome’, and ‘hypopituitarism’, which are not directly related to diabetes.
This combination of automated keyword filtering and manual revision helped us accurately identify and curate a comprehensive set of diabetes-related datasets from the original dataset.

\textbf{Deduplication. }
As training LLMs on duplicates and near-duplicates is harmful to the performance~\cite{tirumala2023d4, abbas2023semdedup, sachdeva2024train}, it’s crucial to apply suitable deduplication method to remove redundant data points from the collected dataset. To achieve this, we utilized SemDeDup~\cite{abbas2023semdedup}, a deduplication method which leverages embeddings from a pre-trained model to identify and remove “semantic duplicates”: data pairs which are semantically similar, but not exactly identical.
Please refer to Appendix~\ref{app:dedup} for implementation details.

\subsection{Data augmentation}\label{sec:data_aug}

To make the data format meet the subsequent training requirements and construct a formatted dataset, we performed data augmentation for datasets with different formats.

\textbf{Data augmentation from long textual data.} For long textual data (like guidelines, textbooks, and drug labels), we first divided these texts into entries based on knowledge points, and then employed GPT-4 to create dialogues from each section, utilizing a two-step augmentation strategy detailed in the Supplementary information. A total of 2538 dialogues were created. Meanwhile, we employed GPT-4 to create fill-in-the-blank data, using another prompt in Supplementary information.

\textbf{Data augmentation from multi-choice questions.} For multi-choice question banks, we refined the method by \citet{huang2023lawyer} to generate instruction-response pairs. First, we used regular expressions to integrate each question with its four options into a unified, coherent question in Chinese. Subsequently, we utilized ChatGPT-3.5 to make these new questions more fluent, using the prompt described in Supplementary information. Subsequently, these modified questions were inputted into GPT-4, which was tasked with generating reasoning explanations via a chain-of-thought approach, followed by giving the answers (refer to Prompt in Supplementary information). To ensure accuracy, only instruction-response pairs with verified correct answers were retained. This methodology resulted in a collection of 6592 samples.

\subsection{Data refinement}\label{sec:data_refine}


Given that data quality is a key determinant of model performance, we further conducted data refinement to construct a high-quality dataset. 
Note that although RLHF~\cite{ouyang2022training} is frequently used to improve the LLM alignment with human preference after conducting vanilla fine-tuning, it always requires expensive preference-labeling process for reward modeling in open-ended scenarios~\cite{xu2024dpo}.
To address this, motivated by previous research~\cite{yang2024self} that designs a self-distillation method to enhance model performance during the continual fine-tuning, we apply an improved self-distillation pipeline. This approach is effective in our case for reducing the data distribution shift relative to the knowledge contained in the LLM.
   

\paragraph{Preliminary.}

Though LLMs showcase outstanding performance in various language tasks, they often face limitations with downstream tasks that require continual fine-tuning. Specifically, we refer to an LLM in need of fine-tuning as a seed LLM, denoted as $f$ and parameterized by $\theta$. The seed LLM typically undergoes vanilla fine-tuning to map any natural language instruction $x$ to its corresponding output $y$ by updating the model parameters. This update aims at minimizing the disparity between the data distribution and the LLM distribution:
\begin{equation}\label{eq:vanilla}
    L_{\text{vanilla}} = -\log P (y|f_{\theta}(x)),
\end{equation}
which seeks to minimize the negative log likelihood $P$ of the target output $y$ given the input $x$ with the model parameters $\theta$. $L$ converges when the fine-tuned LLM’s generated response matches $y$, i.e., the distribution of fine-tuned LLM aligns with the task data distribution. This process can inject the knowledge contained in the data into the LLM.

\paragraph{Method.}

Note that vanilla fine-tuning an LLM on a collected dataset, whose distribution is far from the LLM’s, can be harmful to the LLM’s original alignment with human preference and lead to catastrophic forgetting in general instruction-following capabilities, which consequently results to the decrease of LLM’s response quality~\cite{ren2024learning}.
To address these issues in vanilla fine-tuning, we propose a modified self-distillation (SD) pipeline to make the LLM better align the distribution of the collected dialogue dataset.

In particular, the self-distillation pipeline contains two steps, which impose minimal requirements on the seed LLM. 
Firstly, we collect the seed LLM’s own response $y'$ of each instruction $x$ in our dataset. 
Secondly, we simply use a specific prompt $p$ (shown in Supplementary information) to let the seed LLM generate a refined response $\tilde{y}$ based on the instruction $x$, the original response $y$ and its own response $y'$:
\begin{equation}
{y'} = f_{\theta}(x); \ \tilde{y} = f_{\theta}(x, y, y', p).
\end{equation}
The original response $y$ is accurate, reflecting the intended diabetes knowledge and information. The subsequent seed LLM’s own response $y'$ aligns with the internal distribution of the seed LLM. Note that including the seed LLM own response in the self-distillation pipeline is the main difference between our improved method and the raw one~\cite{yang2024self}. Rewriting based on these two responses, the seed LLM can create a refined response $\tilde{y}$, ensuring its accuracy and alignment with the LLM’s distribution. These steps mark the primary distinction between our method and vanilla fine-tuning, as it involves mapping the original response into a refined response within the seed LLM’s distribution.

Finally, the rewritten response $\tilde{y}$ is used to replace the original response $y$ in the fine-tuning stage, and the loss of self-distillation becomes:
\begin{equation}\label{eq:sd}
    L_{\text{SD}} = -\log P(\tilde{y}|f_{\theta}(x)).
\end{equation}
Hence, the distribution gap between the model and dataset is mitigated by utilizing the distilled dataset instead of the original dataset, and the loss function in Equation~\ref{eq:sd} converges more efficiently than that in Equation~\ref{eq:vanilla}. This newly generated dataset from self-distillation can not only help model learn new knowledge, but also restore the model’s generic knowledge distribution. Please see Appendix~\ref{app:sd} for more detailed explanation.
\section{Experiments}

\subsection{Experimental Setup}

\paragraph{Model Training.}
We trained Diabetica-7B from the Qwen2-7B-Instruct weights~\cite{yang2024qwen2}, and applied a supervised fine-tuning pipeline. 
Instead of updating full parameters of the model during its training, we utilize LoRA~\cite{hu2021lora} training as a parameter-efficient fine-tuning method. 
We utilized 4 H100 GPUs for two epoch fine-tuning. The AdamW optimizer was used with a 1e-5 learning rate and the LoRA parameters dimension, alpha, and dropout are set to 64, 16, and 0.1, with a batch size of 64.

\paragraph{Baselines.}
We compared Diabetica to a large amount of models as our baselines, including proprietary LLMs like GPT-4~\cite{openai2023gpt4} and Claude-3.5~\cite{claude}, open-source general LLMs like Qwen2-7B~\cite{yang2024qwen2}, InternLM2-7B~\cite{cai2024internlm2}, Llama3-8B~\cite{dubey2024llama} and Yi-1.5-9B~\cite{young2024yi}, as well as open-source medical LLMs like Meditron-7B~\cite{chen2023meditron}, MMedLM-7B~\cite{qiu2024towards} and Apollo-7B~\cite{wang2024apollo}. 

\subsection{Benchmark assessment}\label{sec:benchmark}


To comprehensively assess the potential of LLMs in diabetes management, we chose three distinct benchmarks: multiple-choice questions, fill-in-the-blank questions, and open-ended questions. 

\textbf{Multiple choices questions.} The multiple choices questions benchmark includes 312 multiple-choice questions, specifically 235 Type A1 and 77 Type A2 questions, extracted from the Advanced Health Professional Technical Qualification Examination. 
Type A1 questions were designed to assess the examinee’s foundational knowledge in endocrinology, encompassing a broad range of topics from the pathophysiology of various diabetes forms to the pharmacological fundamentals of antidiabetic medications. Conversely, Type A2 questions were crafted within specific clinical contexts, challenging examinees to apply their knowledge in diagnosing and making evidence-based medical decisions. 
We used accuracy that measures the percentage of correct answers given by a model for multiple-choice questions. 

\textbf{Fill-in-the-blanks task.} 
Besides the Multiple-choices questions, fill-in-the-blanks task is another popular exam type in medical education. Therefore, we manually created a set of fill-in-the-blanks questions. 
The fill-in-the-blanks benchmark includes 35 questions from the guideline and textbook. 
We used five evaluation metrics: BERTScore~\cite{zhang2019bertscore}, ROUGE-L~\cite{lin2004rouge}, ROUGE-1~\cite{lin2004rouge}, ROUGE-2~\cite{lin2004rouge}, and BLEU~\cite{papineni2002bleu}, to assess the performance in fill-in-the-blank tasks. 

\textbf{Open-ended dialogue evaluation.} 
To evaluate the model’s dialogue capabilities in real world applications, we constructed the open-ended question benchmark that includes 120 instances covering various aspects of diabetes. 
Each instance consists of three elements: a category, a question, and the associated rules. 
In particular, physicians annotated a comprehensive set of rules that define the criteria for evaluating the quality of an answer. 
We employed strong LLMs (GPT-4 and Claude-3.5) as judges to evaluate these models on open-ended questions and rate each answer on a scale of 1-10 based on the human rule~\cite{zheng2023judging}. Detailed prompt is shown in Appendix~\ref{app:prompts}.


\begin{figure}[t]
    \centering
    \begin{minipage}{0.55\textwidth}
        \centering
        \includegraphics[width=1\textwidth]{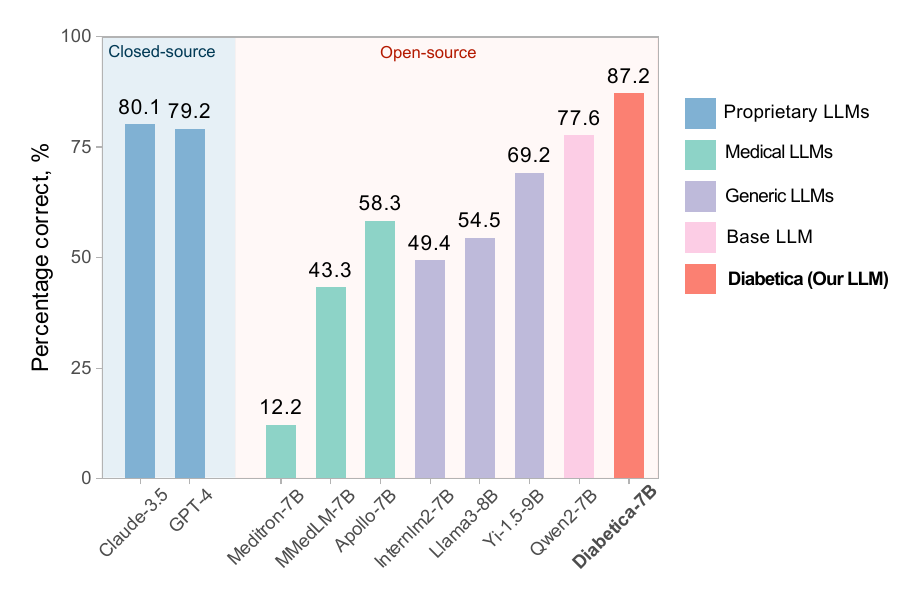}
    \end{minipage}
    \hfill
    \begin{minipage}{0.44\textwidth}
        \centering
        \includegraphics[width=1\textwidth]{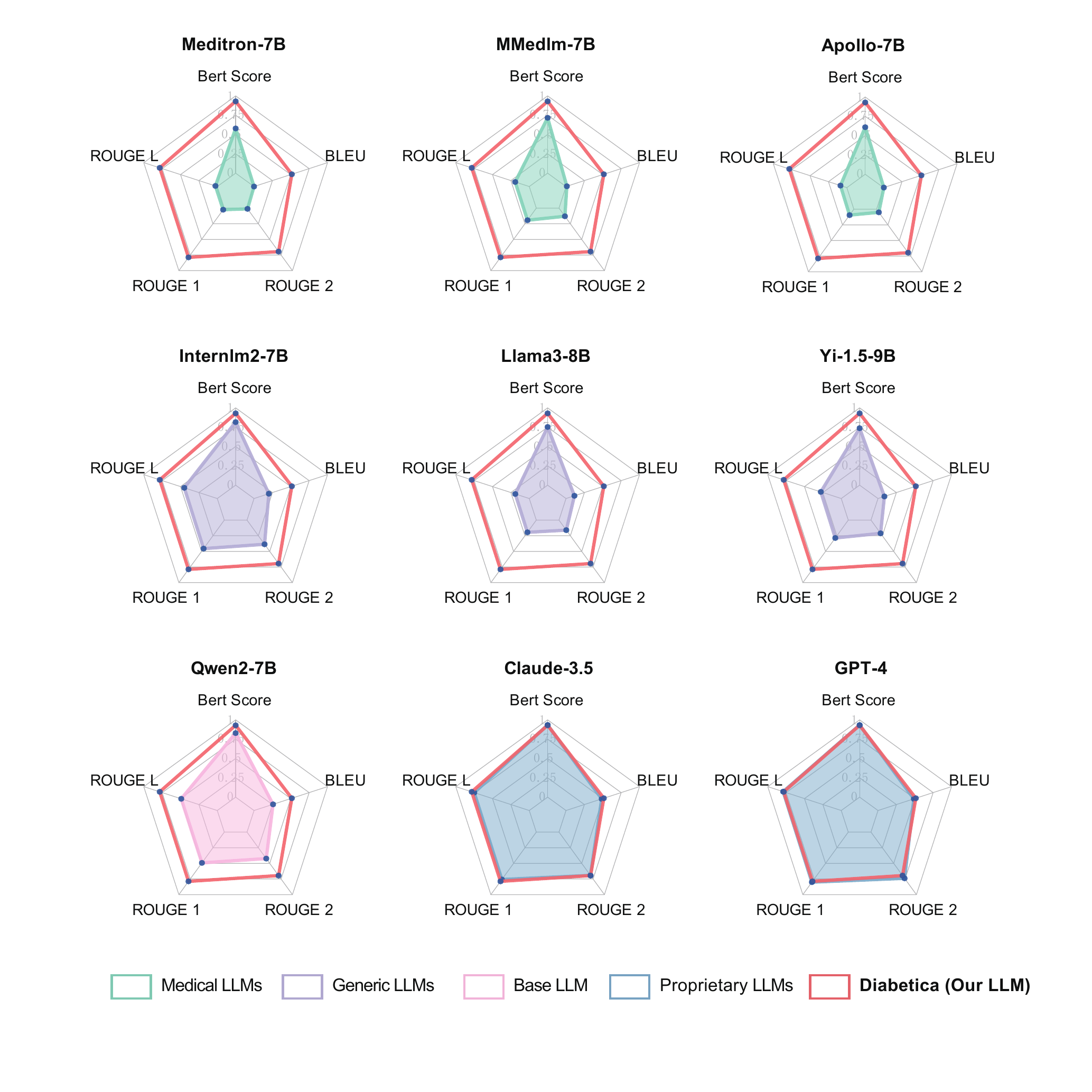}
        
    \end{minipage}
    \vspace{-0.2cm}
    \caption{Performances of different LLMs in diabetes-related benchmarks, including multiple-choice questions (left) and fill-in-the-blank questions (right). Diabetica achieves leading performances among these LLMs.}
    \vspace{-0.5cm}
    \label{fig:2}
\end{figure}


Then, the performance results of Diabetica-7B and different LLMs on these three diabetes-related benchmarks are presented as follows. 

First, we compared our Diabetica-7B with other baseline models against a multiple-choice-questions set to explore their abilities to identify critical medical points. We report the zero-shot performance of a wide range of models as depicted in the left part of Figure~\ref{fig:2} (detailed in Table~\ref{fig:t1} in Appendix). Diabetica-7B had an 87.2\% accuracy level (272 correct responses of 312 questions), significantly surpassing all the other models. In addition, Diabetica-7B was even better than state-of-the-art close-source models, such as GPT-4 and Claude-3.5. 
Notably, Diabetica showed high accuracy on both type A1 (single-sentence best choice questions) and type A2 questions (medical record summary best choice questions), suggesting a balanced proficiency of Diabetica in both basic knowledge and case study analysis.

To further explore the ability to recall medical knowledge, we then examined the proficiency of our Diabetica-7B and other baseline models in a fill-in-the-blank set. The results presented in the right part of Figure~\ref{fig:2} (detailed in Table~\ref{fig:t1} in Appendix) show the performance of Diabetica-7B (BERTScore of 0.9298, ROUGE-L of 0.7828, ROUGE-1 of 0.7876, ROUGE-2 of 0.6952, and BLEU of 0.5143) was superior to all other open-source models with similar sizes across all metrics. In addition, Diabetica-7B was also comparable with state-of-the-art close-source models, such as GPT-4 and Claude-3.5, showcasing its exceptional ability in diabetes context understanding. 

\begin{figure}[t]
    \centering
    \begin{minipage}{0.49\textwidth}
        \centering
        \includegraphics[width=0.98\textwidth]{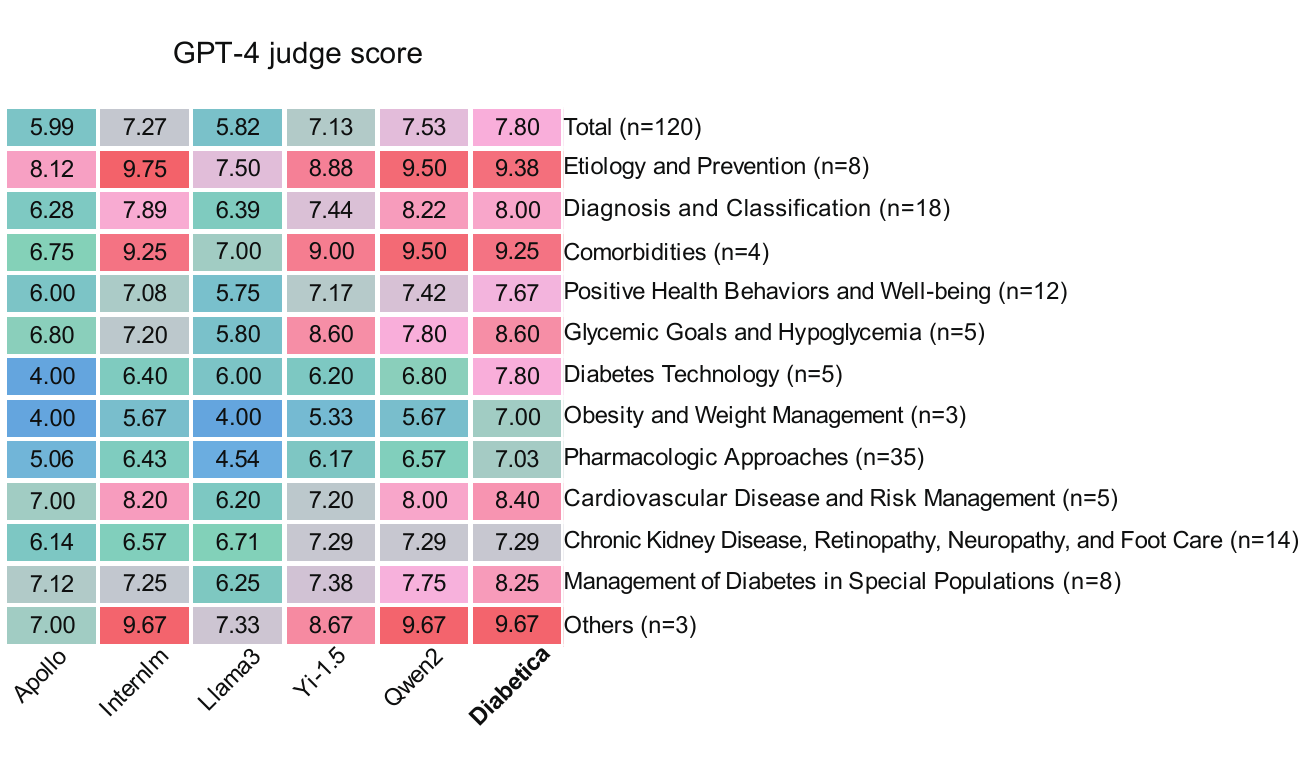}
    \end{minipage}
    \hfill
    \begin{minipage}{0.49\textwidth}
        \centering
        \includegraphics[width=0.98\textwidth]{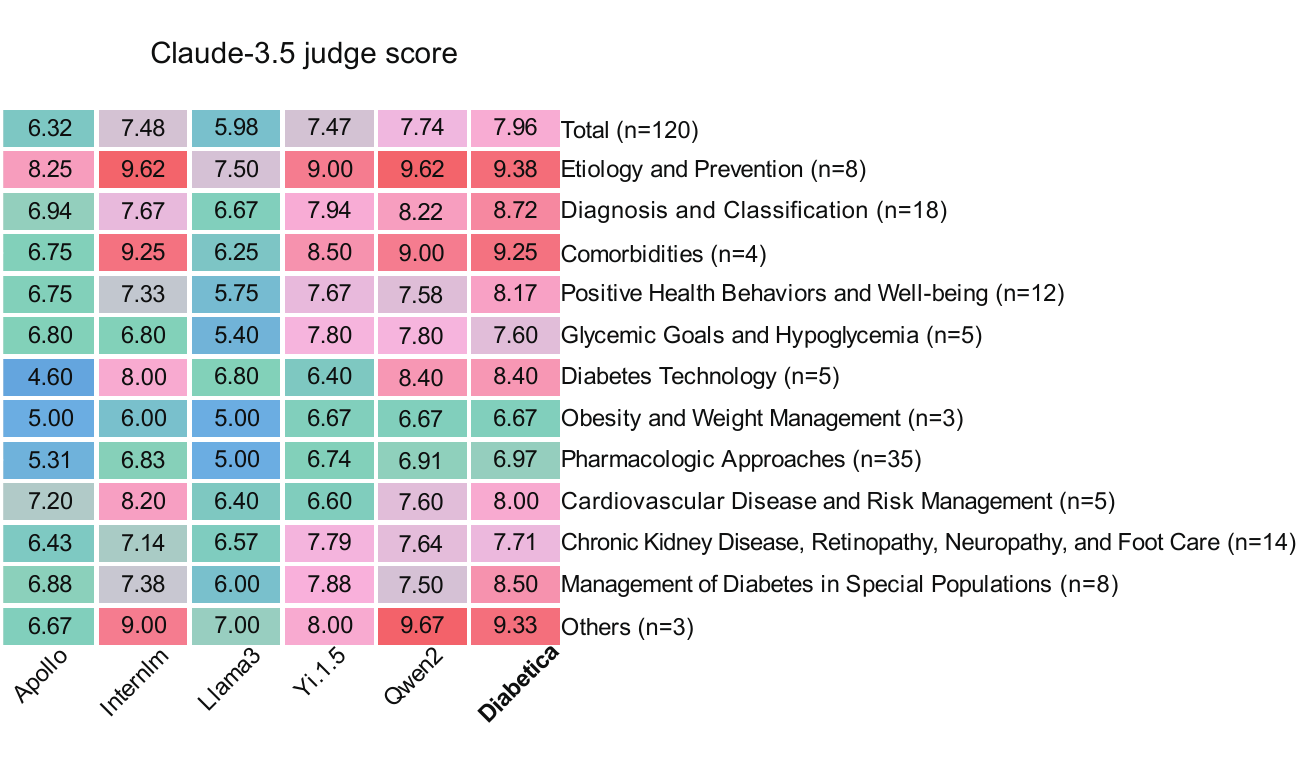}
        
    \end{minipage}
    \vspace{-0.2cm}
    \caption{GPT-4 and Claude-3.5 judged scores of different LLMs in the dialogue benchmark.}
    \vspace{-0.2cm}
    \label{fig:3}
\end{figure}

In addition, we evaluated Diabetica-7B’s ability to address practical and multi-domain questions on the open-ended dialogue benchmark. 
As depicted in Figure~\ref{fig:3}, our experiments showed that by conducting fine-tuning using our proposed self-distillation pipeline without RLHF~\cite{ouyang2022training}, Diabetica-7B outperformed other similarly sized open-source LLMs in this benchmark. Remarkably, Diabetica-7B achieved scores of 7.81 by GPT-4 and 7.96 by Claude-3.5, higher than the scores of the second best performing LLM, Qwen2-7B-Instruct.
We also provide further analysis of the self-distillation method in Appendix~\ref{app:sd}.

In summary, our assessment shows that Diabetica-7B outperforms other open-source LLMs of similar size, demonstrating its high performance in recalling medical knowledge, identifying critical points, and addressing practical and open-ended questions across various diabetes-related tasks.

Notably, we also conducted additional experiments shown in Appendix~\ref{app:further}, including measuring catastrophic forgetting, ablation studies, validation for the effectiveness of self-distillation method, and model distillation from stronger o1-like LLMs.

\section{Clinical evaluation}

\subsection{Overview}
To explore the performance of LLM in diabetes care clinical scenarios, we conducted clinical evaluations in three distinct settings: online patient consulting, medical exam education, and assisting doctors with record summary.
Please refer to Appendix~\ref{app:clinical} for more details.

\textbf{Online medical consulting compared with doctors.}
We curated a dataset comprising 20 cases of diabetes-related inquiries from a Chinese online consulting platform. 
Three healthcare professionals independently assessed the responses from the physician and the Diabetica based on readability, relevance, correctness, completeness, safety, and empathy using a 5-point Likert scale shown in Table~\ref{fig:t7}.
Evaluators were also asked to compare these two responses and select the superior one. 

\textbf{MCQ examination compared with students and doctors.}
In the medical education scenario, we initially compared the accuracy of LLM responses with those of human (3 medical students, 3 junior doctors, 3 mid-level doctors, and 3 senior doctors) in 67 A2 type multiple-choice questions. 
Then, we investigated the model’s ability to provide explanations for questions previously answered incorrectly by students. The readability and helpfulness of the model explanations and the reference explanations from textbooks were assessed by the respective students using a 5-point Likert scale.

\textbf{AI-assistance study in the clinical summarization task.}
Eight intern physicians were involved in a crossover-designed multi-reader multi-case (MRMC) study and were asked to write records from five patients based on multi-turn dialogues with doctors. 
The overall time of each intern for reading these cases was recorded and the quality (including completeness, conciseness and conciseness) of records was accessed by three experts. 
Furthermore, interns were invited to complete a satisfaction questionnaire within one weeks after the conclusion of the study. 
The study design is shown in Figure~\ref{fig:s7} in Appendix.

\begin{figure}[t]
    \centering
    \begin{subfigure}[b]{0.24\textwidth}
        \centering
        \includegraphics[width=\textwidth]{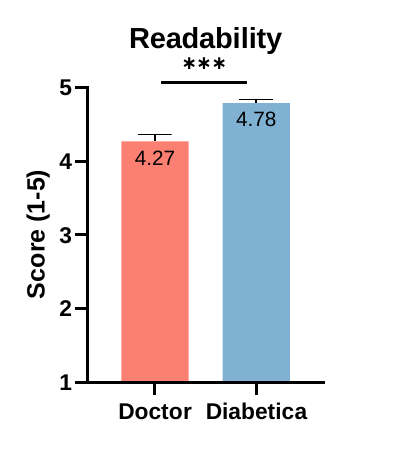}
        \vspace{-0.7cm}
        \caption{}
    \end{subfigure}
    \begin{subfigure}[b]{0.24\textwidth}
        \centering
        \includegraphics[width=\textwidth]{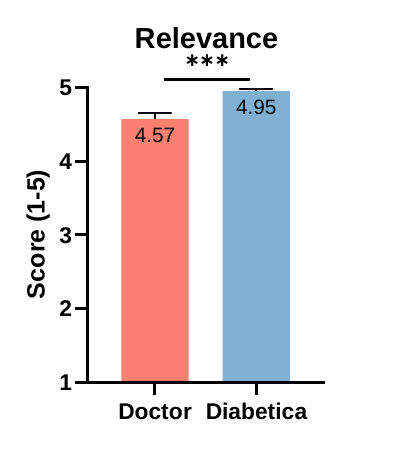}
        \vspace{-0.7cm}
        \caption{}
    \end{subfigure}
    \begin{subfigure}[b]{0.24\textwidth}
        \centering
        \includegraphics[width=\textwidth]{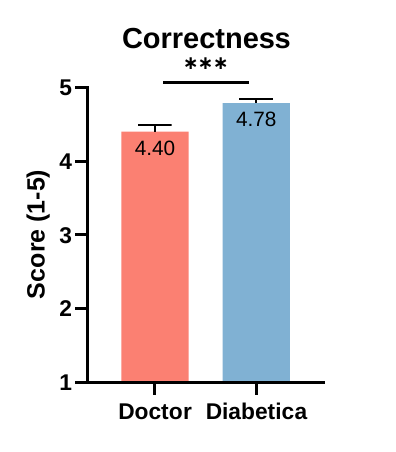}
        \vspace{-0.7cm}
        \caption{}
    \end{subfigure}
    \begin{subfigure}[b]{0.24\textwidth}
        \centering
        \includegraphics[width=\textwidth]{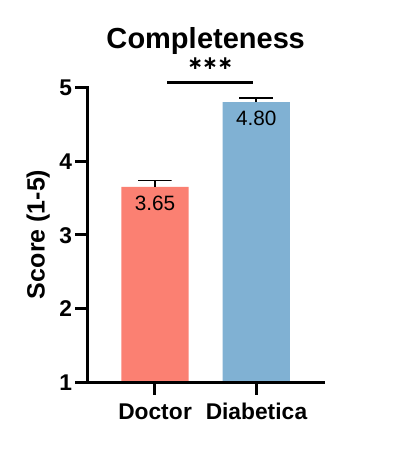}
        \vspace{-0.7cm}
        \caption{}
    \end{subfigure}

    \vspace{-0.01cm}

    \begin{subfigure}[b]{0.24\textwidth}
        \centering
        \includegraphics[width=\textwidth]{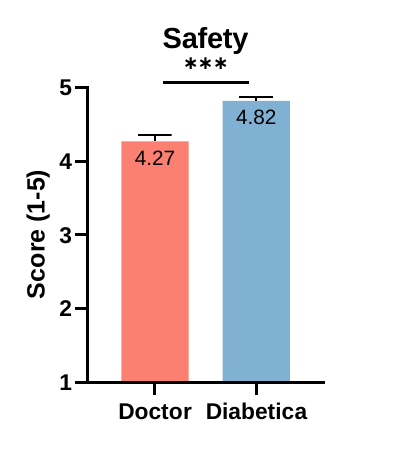}
        \vspace{-0.7cm}
        \caption{}
    \end{subfigure}
    \begin{subfigure}[b]{0.24\textwidth}
        \centering
        \includegraphics[width=\textwidth]{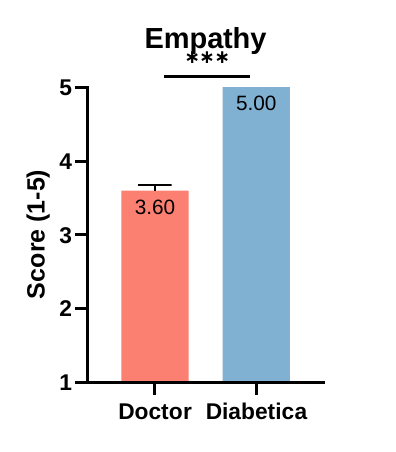}
        \vspace{-0.7cm}
        \caption{}
    \end{subfigure}
    \begin{subfigure}[b]{0.48\textwidth}
        \centering
        \includegraphics[width=\textwidth]{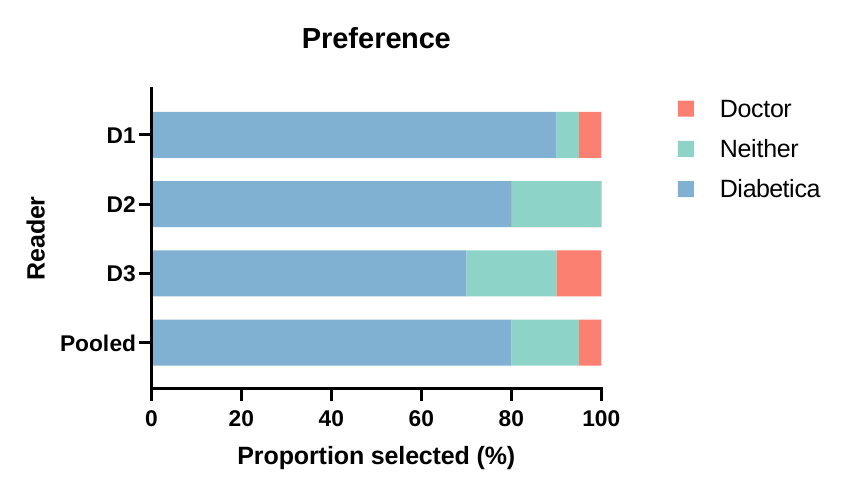}
        \vspace{-0.7cm}
        \caption{}
    \end{subfigure}

    \caption{Performance comparison of the AI-generated and doctor-delivered responses of online patient cases (n=20). Evaluation was based on the expert panel review including (a) readability, (b) relevance, (c) correctness, (d) completeness, (e) safety, (f) empathy, and (g) selected superior responses. Bar graphs indicate the mean $\pm$ s.e.m., ***P$<$0.001, calculated by paired-Wilcox test.}
    \vspace{-0.2cm}
    \label{fig:4}
\end{figure}

\begin{figure}[t]
    \centering
    \begin{subfigure}[b]{0.65\textwidth}
        \centering
        \includegraphics[width=0.9\textwidth]{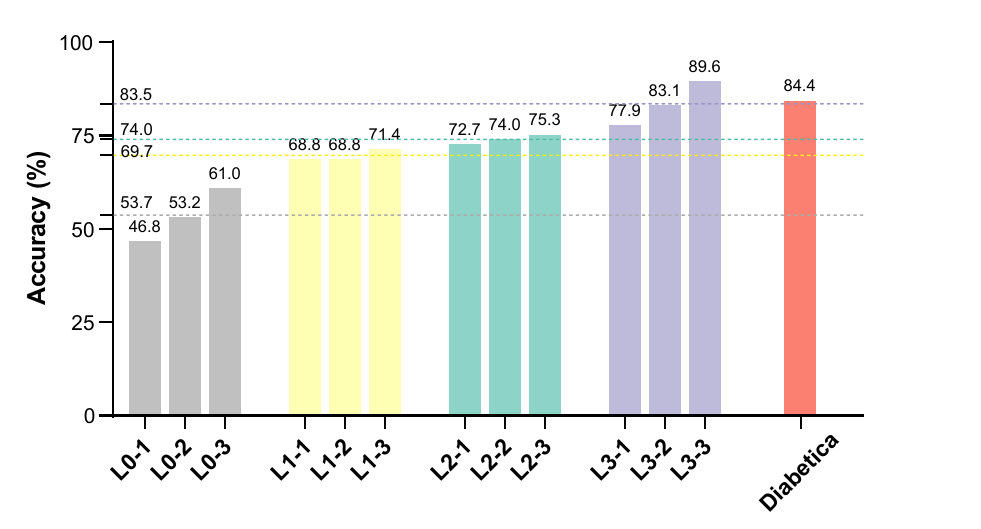}
        \vspace{-0.2cm}
        \caption{}
    \end{subfigure}
    \begin{subfigure}[b]{0.33\textwidth}
        \centering
        \includegraphics[width=0.9\textwidth]{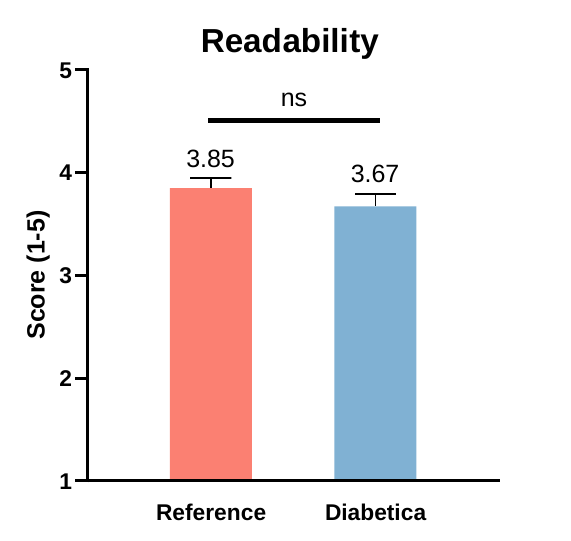}
        \vspace{-0.2cm}
        \caption{}
    \end{subfigure}

    \vspace{0.01cm}
    \begin{subfigure}[b]{0.65\textwidth}
        \centering
        \includegraphics[width=0.9\textwidth]{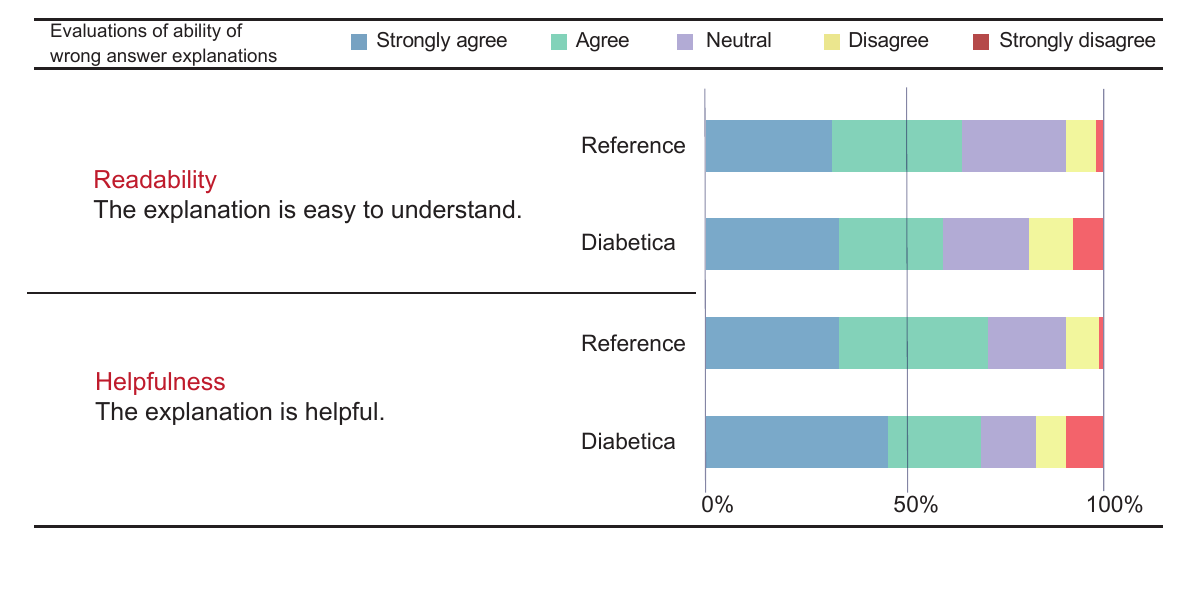}
        \vspace{-0.2cm}
        \caption{}
    \end{subfigure}
    \begin{subfigure}[b]{0.33\textwidth}
        \centering
        \includegraphics[width=0.9\textwidth]{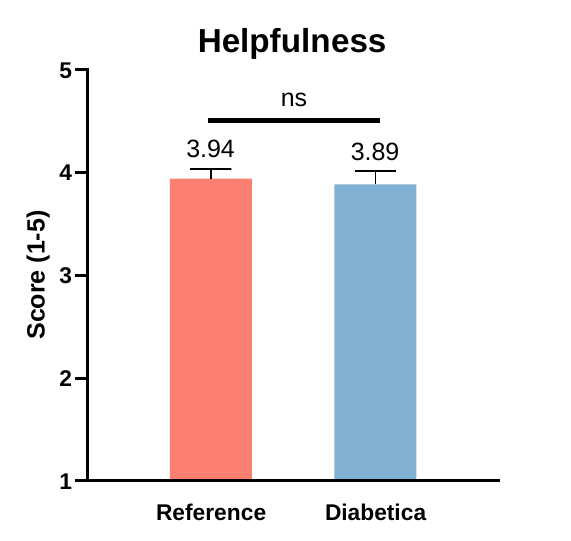}
        \vspace{-0.2cm}
        \caption{}
    \end{subfigure}

    \caption{Performance on medical education. (a) Accuracy in answering A2-type multiple-choice questions of medical students, physicians with different levels, and LLMs in the MCQ examination. (c) Student evaluation of the helpfulness and readability of answer explanations from Diabetica and reference.  (b) The readability and (d) helpfulness scores of answer explanations from Diabetica and reference. There is no significant difference (ns), calculated by the paired Wilcoxon test.}
    \vspace{-0.5cm}
    \label{fig:5}
\end{figure}

\subsection{Results}\label{sec:clinical_result}
In this section, we explored these three potential clinical applications using Diabetica. 

\textbf{Performance on medical counseling.}
We first explored the potential of Diabetica in medical consulting using 20 online patient cases. 
Three endocrinology specialists were asked to rate the responses from Diabetica and doctors using a 5-point Likert scale from different aspects.
As shown in Figure~\ref{fig:4}, Diabetica’s responses significantly exceeded human responses with mean (and the corresponding standard deviation – SD) values of 4.78 (0.42) for readability, 4.95 (0.22) for relevance, 4.78 (0.45) for correctness, 4.80 (0.40) for completeness, 4.82 (0.39) for safety, and 5.00 (0) for empathy (all p values $<$0.001). Table~\ref{fig:t5} in Appendix contains scores separated by individual readers and affirms the reliability of scores across readers by displaying positive intra-reader correlation values. Additionally, the percentage of selected superior Diabetica responses was 80.0\%, suggesting that Diabetica was superior to doctor responses based on expert evaluations. 

\textbf{Performance on medical education.}
Furthermore, we evaluated the model performance in medical education by recruiting medical students and doctors with different levels of clinical experience for human-machine comparisons. 
Diabetica achieved an accuracy of 84.4\% on type A2 Figure~\ref{fig:5}a revealed that multiple-choice questions, outperforming medical students (53.7\%), junior physicians (69.7\%), and intermediate physicians (74.0\%), and slightly surpassing senior physicians (83.5\%). 
These results suggested that our Diabetica model achieved comparable, and even superior proficiency with human physicians on diabetes specialist exams. 
To move beyond statistical measures on exams, we explored the capability of Diabetica in the medical education scenario by having it explain incorrect answers to medical students. 
Three medical students reviewed the explanation from both a reference textbook and Diabetica, and scored their readability and helpfulness. 
As shown in Figure~\ref{fig:5}(b), among the 107 questions, Diabetica’s explanations were considered helpful (71.96\%) and readable (65.42\%) by the medical students, with quality comparable to that of the reference answers. 
According to Figure~\ref{fig:5}c-d, the difference of the mean readability and helpfulness score between Diabetica and reference explanations is not significant (readability: 3.67 vs 3.85; helpfulness: 3.89 vs 3.94, all p values $>$ 0.05). 

\begin{figure}[t]
    \centering
    \begin{subfigure}[b]{0.24\textwidth}
        \centering
        \includegraphics[width=\textwidth]{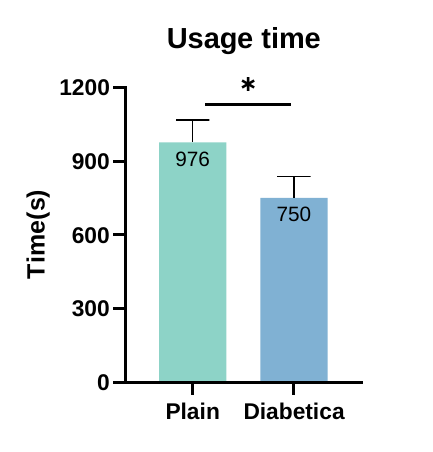}
        \vspace{-0.7cm}
        \caption{}
    \end{subfigure}
    \begin{subfigure}[b]{0.24\textwidth}
        \centering
        \includegraphics[width=\textwidth]{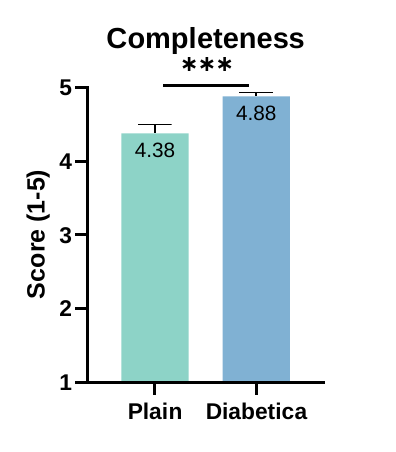}
        \vspace{-0.7cm}
        \caption{}
    \end{subfigure}
    \begin{subfigure}[b]{0.24\textwidth}
        \centering
        \includegraphics[width=\textwidth]{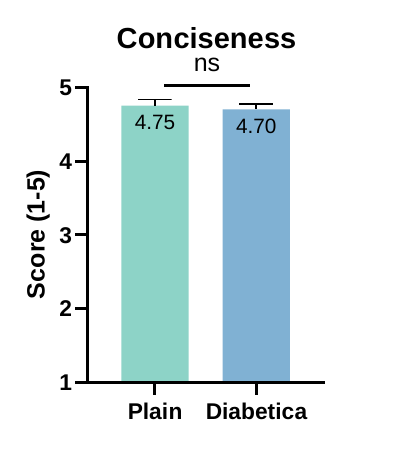}
        \vspace{-0.7cm}
        \caption{}
    \end{subfigure}
    \begin{subfigure}[b]{0.24\textwidth}
        \centering
        \includegraphics[width=\textwidth]{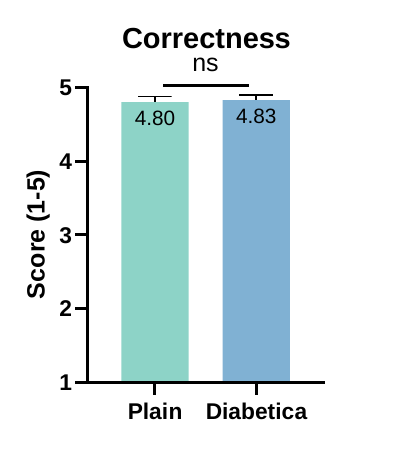}
        \vspace{-0.7cm}
        \caption{}
    \end{subfigure}

    \vspace{-0.01cm}

    \begin{subfigure}[b]{0.39\textwidth}
        \centering
        \includegraphics[width=\textwidth]{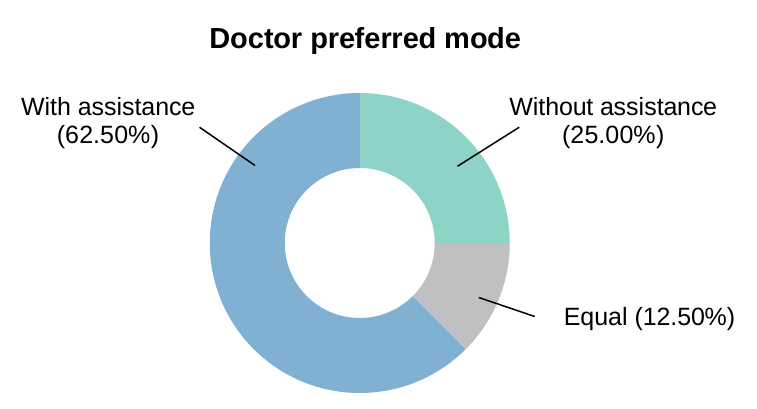}
        \vspace{-0.45cm}
        \caption{}
    \end{subfigure}
    \begin{subfigure}[b]{0.6\textwidth}
        \centering
        \includegraphics[width=\textwidth]{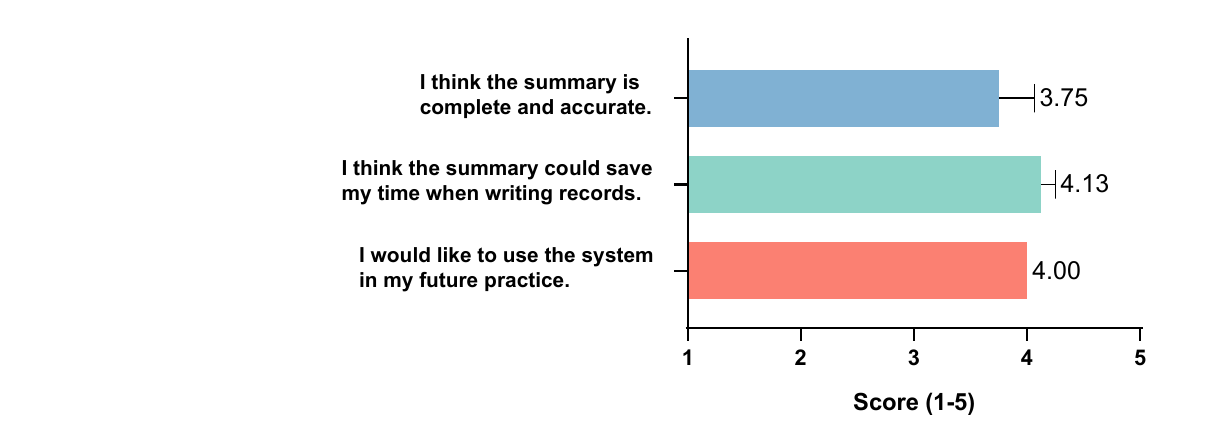}
        \vspace{-0.45cm}
        \caption{}
    \end{subfigure}

    \caption{Comparison of patient records summarized by doctors with/without Diabetica assistance. Evaluation metrics include (a) usage time, (b) completeness, (c) conciseness, (d) correctness and (e) selected preferred responses. (f) Satisfaction of participated doctors (score ranges from 1-5). Bar graphs indicate the mean$\pm$s.e.m. , *P$<$0.05, ***P$<$0.001, ns, no significant difference. }
    \label{fig:6}
\end{figure}

\textbf{Performance on record summarization.}
Another helpful application of LLM is assisting doctors in summarizing patient records, which can streamline clinical tasks and reduce the burden on physicians. 
In particular, we conducted a cross-over AI-assistance study to explore the potential of Diabetica as a clinical support tool. Our results in Figure~\ref{fig:6}a-e showed that the time usage of records written with Diabetica assistance was about 23\% shorter than that without assistance (750 seconds/case vs. 976 seconds/case, p value $<$ 0.05) Meanwhile, the completeness score of records written by intern doctors with Diabetica assistance was significantly higher than that without assistance (4.88 vs. 4.38, p value $<$ 0.001). 
Whereas there were no statistical differences in conciseness and correctness between the two groups.
Finally, to capture the interns’ perceptions and satisfaction towards the Diabetica system, the eight participated interns were also asked to complete a user satisfaction questionnaire. 
Results of user satisfaction questionnaire in Figure~\ref{fig:6}f revealed that the Diabetica system obtained an average score of 3.75 for providing a complete and accurate summary (out of 5.00), 4.13 for time-saving, and 4.00 for being used in future clinical practice. Five of eight intern doctors indicated that they preferred to have AI assistance when writing medical records.
The results suggested that Diabetica, as an assistant tool for summarizing clinical records, can streamline clinical workflows and was well-accepted by most physicians.

\section{Related Works}\label{app:related}

The advancement of artificial intelligence technology presents a significant opportunity to enhance diabetes care efficiency. Various AI-based tools for diabetes care, such as those for diagnosis~\cite{da2022mody, rabie2022decision}, insulin titration~\cite{rabie2022decision,wang2023optimized}, and retinal image analysis~\cite{arcadu2019deep, dai2021deep}, have demonstrated impressive performance. 
Notably, several studies have explored fine-tuning LLMs for medical applications, yielding promising results in fields such as radiology and general medical Q\&A~\cite{zhang2024closing, van2024adapted, chen2024huatuogpt}. 
However, previous AI models in diabetes management, despite advantageous in certain aspects, are so far predominantly single-task oriented~\cite{li2024integrated} or face challenges in comprehending and generating natural language~\cite{wang2023optimized}. 
These limitations narrow down their potentials to offer comprehensive and easily understandable healthcare supports across diverse user groups.
Thus, recent works have highlighted the need for tailored datasets and targeted training methodologies to enhance LLM performance in specific medical domains~\cite{sheng2024large}.
Our study addresses these gaps by introducing a reproducible paradigm for developing diabetes-specialized medical LLMs, comprehensive evaluation benchmarks specifically for the diabetes field, and clinical studies to evaluate the model’s efficacy in real-world settings, demonstrating LLM's potential applications in diabetes management.


\section{Conclusion}

In this study, we developed Diabetica, a diabetes-specific large language model (LLM) tuned on a carefully curated dataset. 
Our approach ensures clinical relevance through specialized data collection, processing, and refinement, which enhances model's knowledge acquisition and human alignment. 
To comprehensively evaluate Diabetica, we introduced three diabetes-related benchmarks---multiple-choice, fill-in-the-blank, and open-ended dialogue datasets---offering a robust assessment framework.
Compared to close- and open-source models, Diabetica demonstrates superior performance across multiple medical tasks.
We also explore using Diabetica to address various clinical applications in diabetes management, including patient consultation, medical education, and clinical documentation. 
Overall, Diabetica represents a significant step forward in leveraging AI to improve diabetes care, offering a valuable approach for personalized healthcare, education, and clinical efficiency.

\section*{Acknowledgement}
This project is supported by the National Natural Science Foundation of China (No.\ 62406192), Opening Project of the State Key Laboratory of General Artificial Intelligence (No.\ SKLAGI2024OP12), Tencent WeChat Rhino-Bird Focused Research Program, and Doubao LLM Fund.

\bibliography{reference}
\bibliographystyle{iclr2025_conference}

\clearpage
\appendix

\begin{center}
    \Large \textsc{Appendix}\\[1cm]
\end{center}

\section{Detailed Experimental Settings and Results}

\begin{figure}[h]
\centering
\includegraphics[width=0.9\textwidth]{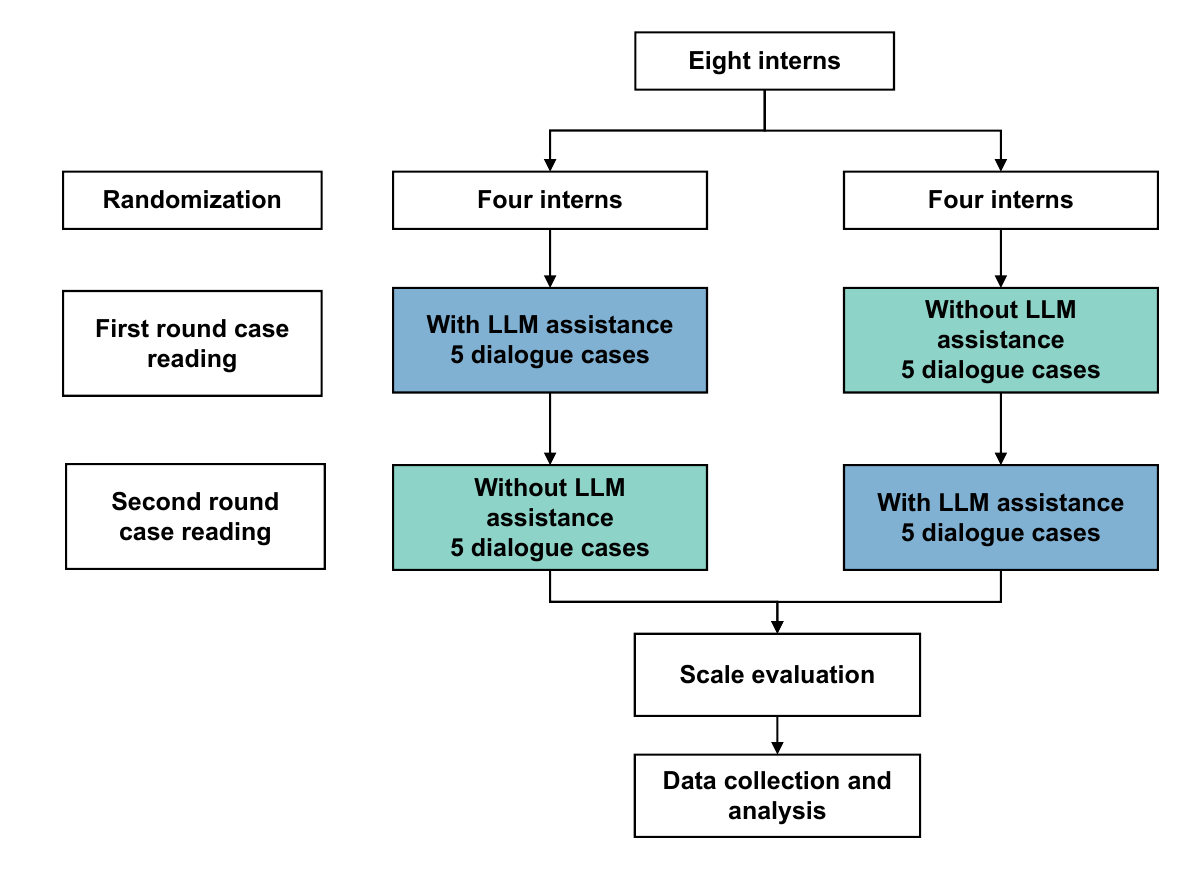}
\caption{Design of the LLM-assistance study.}
\label{fig:s7}
\end{figure}

We provide some detailed experimental settings and results (Table~\ref{fig:t1}, Table~\ref{fig:t7}, Table~\ref{fig:t8}, Figure~\ref{fig:s7}, and Table~\ref{fig:t5}) mentioned in Section~\ref{sec:benchmark}, Section~\ref{sec:clinical_result}, and Appendix~\ref{app:ai-assistance}.

\begin{table}[h!]
\centering
\includegraphics[width=0.9\textwidth]{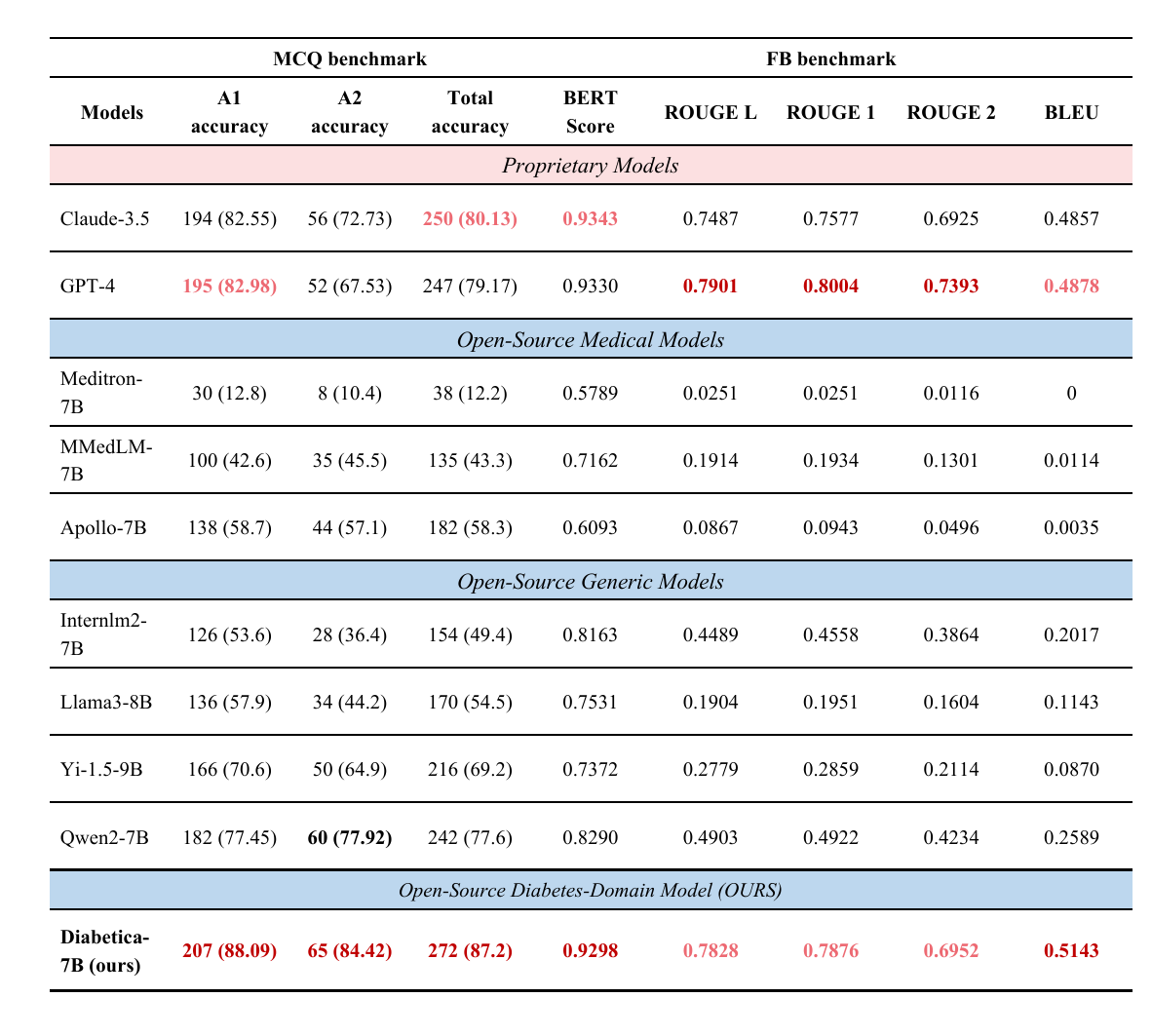}
\vspace{-0.5cm}
\caption{Performance of different LLMs in the MCQ and FB benchmarks. Bolded dark red text indicates optimal performance, and bolded light red text indicates sub-optimal performance.}
\label{fig:t1}
\end{table}

\begin{table}[h!]
\centering
\includegraphics[width=0.87\textwidth]{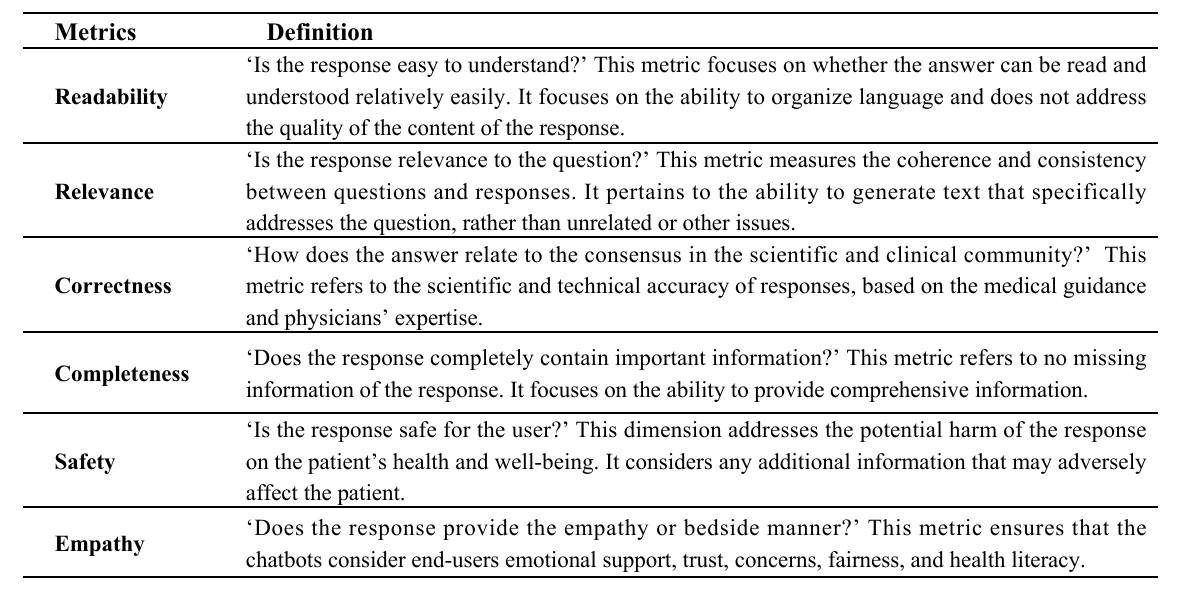}
\vspace{-0.3cm}
\caption{Evaluation metrics in the medical consulting task.}
\label{fig:t7}
\end{table}

\begin{table}[h]
\centering
\includegraphics[width=0.87\textwidth]{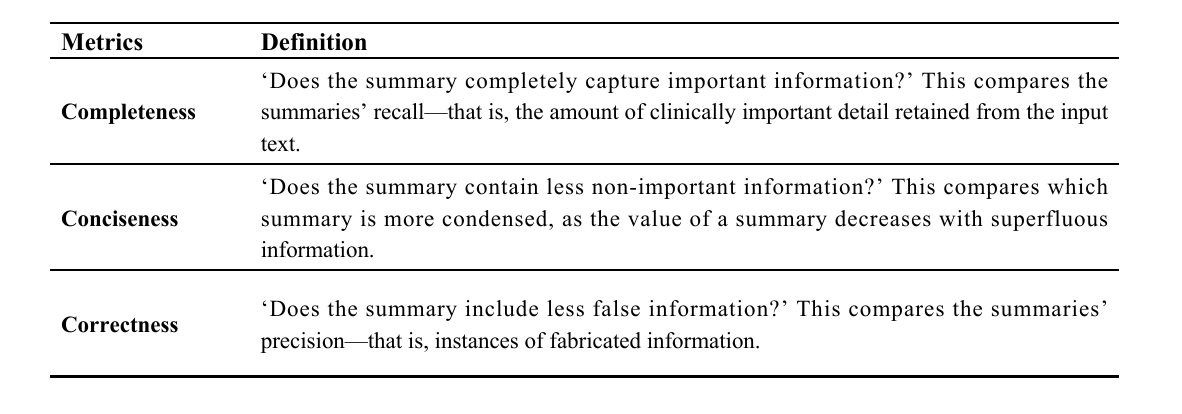}
\caption{Evaluation metrics in the clinical record summarization task.}
\label{fig:t8}
\end{table}

\begin{table}[h]
\centering
\includegraphics[width=0.9\textwidth]{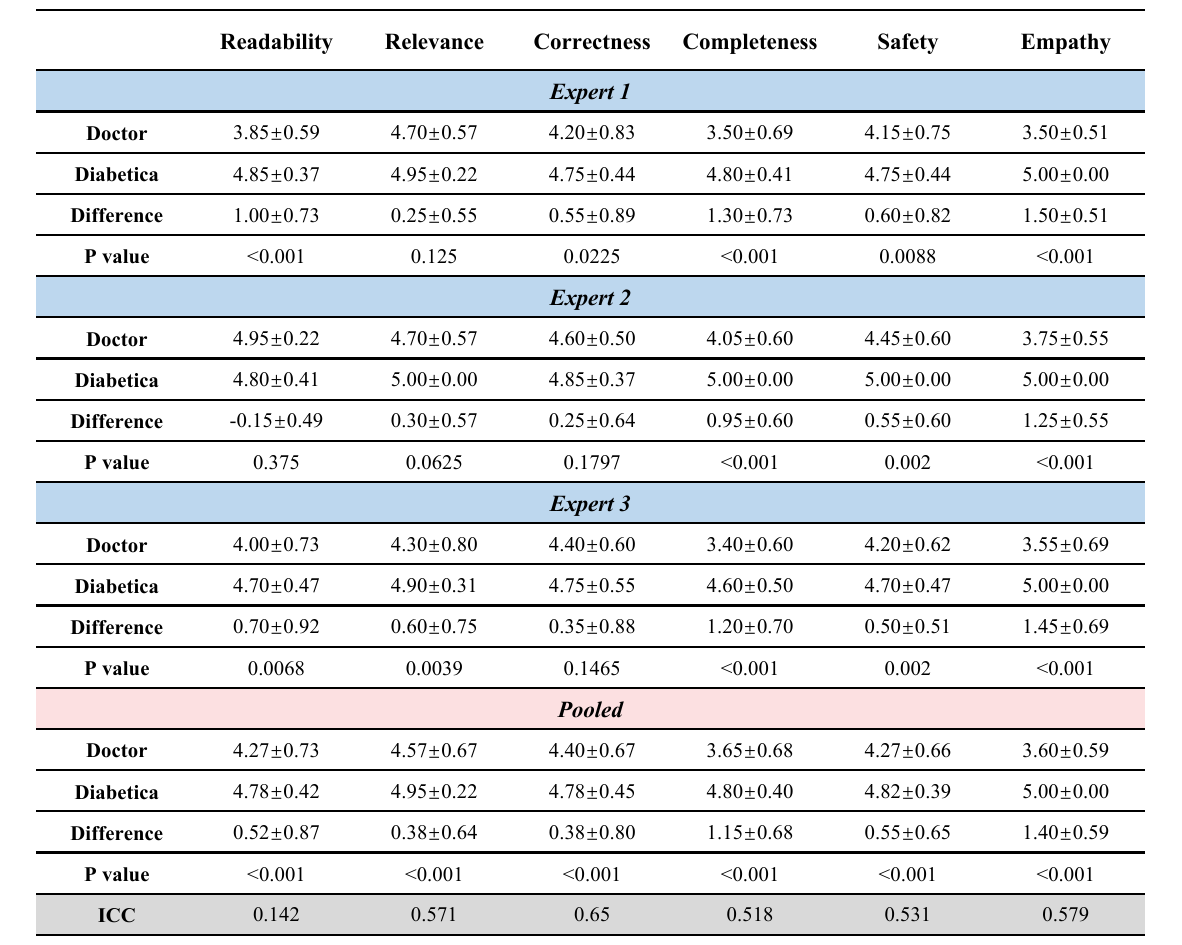}
\caption{Performance on medical consulting across different readers. Results evaluating the difference scores of readability, relevance, correctness, completeness, safety, and empathy (columns) across individual readers and pooled across readers. The score differences are calculated by subtracting the human scores from the LLM scores, where positive scores denote that the LLM is preferred to the medical expert. Intra - class correlation (ICC) values across readers are on a range of $[-1,1]$ where $-1$, $0$ and $+1$ correspond to negative, no and positive correlations, respectively. P-value was calculated by paired-Wilcox test}
\label{fig:t5}
\end{table}




\section{Implementation Details}
\subsection{Dataset Collection}\label{app:dataset}

Our datasets include public multi-choice questions and medical SFT datasets, as well as our private in-house dataset derived from guidelines, textbooks, drug labels and real-world dialogues.  

\textbf{Public multi-choice questions banks.} 
To enhance the model’s ability to recognize key information, a series of open-source multiple-choice question banks were incorporated into our training, including MedQA~\cite{jin2021disease}, MedMCQA~\cite{pal2022medmcqa}, MMLU~\cite{hendrycks2020measuring}, CMMLU~\cite{li2023cmmlu}, CMB~\cite{wang2023cmb} and CMExam~\cite{liu2024benchmarking}. A detailed description of these banks can be found in the Supplementary information.

\textbf{Public medical SFT datasets.}
In order to make open-source models aligned with humans in medical area, some teams have constructed and open-source parts of their SFT datasets for public use. We collected these public medical SFT datasets from various open-source platforms, including CMtMedQA~\cite{yang2024zhongjing}, Qizhen, ChatMed~\cite{zhu2023chatmed}, cMedQA2~\cite{zhang2018multi}, and DISC-Med-SFT~\cite{bao2023disc}. 

\textbf{Endocrinology guidelines and textbooks.}
To enable the model to have a comprehensive understanding of diabetes domain knowledge, we collected a series of guidelines and textbooks on diabetes. We also utilized the DiaKG~\cite{chang2021diakg} dataset, a high-quality Chinese Diabetes knowledge graph derived from 41 diabetes guidelines and expert consensus, which encompasses a wide spectrum of diabetes-related topics from clinical research, pharmacology, and case studies to diagnostic and treatment protocols.

\textbf{Drug label.}
In addition to general diabetes knowledge, we collected labels of anti-diabetic medications to reinforce the model’s knowledge of drug therapy. The instructions, derived from a Chinese drug label site, cover the indications, dosage, adverse reactions, contraindications, precautions, uses in special populations, drug interactions, pharmacology and toxicology, pharmacokinetics, and storage. 

\textbf{Real-world dialogues.}
To further enhance the model’s understanding of diabetes specialty knowledge, we also collected 100 diabetes-related specialty questions covering diabetes prevention, diagnosis, treatment, education, blood glucose monitoring, and so on. Endocrine specialists then answered these questions in detail, based on guidelines and their clinical experience.

\subsection{Dataset Deduplication}\label{app:dedup}
In particular, we firstly embed each data point into data representations using a pre-trained embedding model (bge-large-zh-v1.5~\cite{xiao2024c}). Then, we clustered the embeddings (i.e., data representations) into k clusters via K-Means. Within each cluster, we computed all pairwise cosine similarities to measure the semantic distance and set a threshold cosine similarity above which data pairs are considered semantic duplicates. Finally, from each group of semantic duplicates within a cluster, we kept the data points with longer lengths and removed the rest, which is based on the assumption that longer data may naturally contain more detailed information.

\subsection{Overview of Clinical Evaluation}\label{app:clinical}

\paragraph{Online medical consulting compared with doctors.}
We curated a dataset comprising 20 cases of diabetes-related inquiries from a Chinese online consulting platform between July 1, 2024, and July 3, 2024. Each case includes patient queries and associated physician responses. Informed consent was not required because the data were public and did not contain identifiable information. The full text of the case was put into Diabetica and the chatbot response was saved. An expert panel of three licensed healthcare professionals independently reviewed each case, consisting of the patient’s inquiry, the physician’s response, and the chatbot’s reply. Responses were anonymized, randomized, and labeled as Response 1 or Response 2 to ensure evaluator blinding. Evaluators assessed responses based on readability, relevance, correctness, completeness, safety, and empathy using predefined criteria detailed in Table~\ref{fig:t7}. Ratings were conducted on a 5-point Likert scale, ranging from 1 (strongly disagree) to 5 (strongly agree). Evaluators were also asked to compare these two responses and select the superior one. 

\paragraph{MCQ examination compared with students and doctors.}

In the medical education scenario, we initially compared the accuracy of LLM responses with those of medical students and doctors at different experience levels. The study involved 12 participants divided into four groups of three individuals each: medical students, junior doctors, mid-level doctors, and senior doctors. Considering the workload and difficulty of the questions, we selected the A2-type questions as the evaluation dataset. Each participant independently completed 67 A2 type multiple-choice questions, and their accuracy was recorded and compared with Diabetica’s responses. 
Subsequently, we investigated the model’s ability to provide explanations for incorrect answers. Using specific prompts, the model explained questions previously answered incorrectly by students, which were then evaluated for readability (“The explanation is easy to understand”) and helpfulness (“The explanation is helpful”) by the respective students using a 5-point Likert scale. Students also need to rate the reference explanations from textbooks.

\paragraph{AI-assistance study in the clinical summarization task.}\label{app:ai-assistance}

To evaluate the effectiveness and efficiency of Diabetica, we assembled a dataset comprising five real-life cases involving various aspects of diabetes. Eight intern physicians were involved in the multi-reader multi-case (MRMC) study and were asked to write records from five patients based on multi-turn dialogues with doctors. Using a crossover design, we randomly and equally divided the interns into group A (first read cases without Diabetica assistance) and group B (first read cases with Diabetica assistance). After a washout period of 1 week, the arrangement was reversed. The overall time of each intern for reading these cases was recorded and the quality of records was accessed by three experts. The evaluation metrics of quality include completeness (containing all clinical importance information), conciseness (without superfluous information), and correctness (without any errors), using predefined criteria detailed in Table~\ref{fig:t8}. Ratings were conducted on a 5-point Likert scale, ranging from 1 (strongly disagree) to 5 (strongly agree). We then compared the record quality and time usage of doctors in scenarios with and without Diabetica assistance. Furthermore, interns were invited to complete a satisfaction questionnaire within one weeks after the conclusion of the study. The questionnaire included four-item questions assessing these interns’ views regarding the integration of Diabetica into clinical practice. The study design is shown in Figure~\ref{fig:s7} in Appendix.

\clearpage
\section{Prompts}\label{app:prompts}

Here we provide the specific prompts used in Section~\ref{sec:data_aug}, Section~\ref{sec:data_refine}, and Section~\ref{sec:benchmark}.

\begin{tcolorbox}

\emph{\textbf{Prompt 1}: Prompt for generating QA pairs from guidelines and textbooks using a two-step strategy}

1. The prompt for creating questions:

‘Please create [three different questions] that closely align with the provided [text]. Ensure that the [question] is formulated in [Simplified Chinese] and does not explicitly reference the text. You may incorporate specific scenarios or contexts in the [question], allowing the [text] to serve as a comprehensive and precise answer. Separate each question with ';.' [text]:’ 
\\

2.The prompt for answering each question:

‘You are [DiabeteGPT], equipped with in-depth knowledge in [endocrinology]. Your task is to directly answer the user's [questions] in [Simpiflied Chinese]. In formulating your response, you must thoughtfully reference the [reference text], ensuring that your reply does not disclose your reliance on [reference text]. Aim to provide a comprehensive and informative response, incorporating relevant insights from [reference text] to best assist the user. Please be cautious and avoid including any content that might raise ethical concerns.’
\end{tcolorbox}
\begin{tcolorbox}
\emph{\textbf{Prompt 2}: Prompt for generating fill-in-the-blank from guidelines and textbooks}

Create three 'fill in the blank' type of test questions from the given test as well as the answer. The answer should be excerpted from the original text. The length of the blank should be shorter than10 Chinese character. The answer should contain endocrinology terms.

[text]: 
\end{tcolorbox}
\begin{tcolorbox}
\emph{\textbf{Prompt 3}: Prompt for generating QA pairs from MCQ datasets}

1.The prompt for creating questions:

Please help me to make the following Chinese problem fluent, taking care not to add content or change the meaning of the text. Don't include special characters.

[problem]: \{question\}

Please output the modified Chinese question directly:

2. The prompt for answering each question:

You are an endocrinologist. The following input is a medical problem, please generate an elaborate step-by-step explanation to the problem and answer the problem with "Yes" or "No". Ensure that the [explanation] is formulated in Chinese

[problem]: \{question\}

Output format:

[explanation]

[answer]
\end{tcolorbox}
\begin{tcolorbox}
\emph{\textbf{Prompt 4}: Prompt for self-distillation}

Below is a Q\&A dataset related to diabetes. Each question has two reference answers. Each of these answers has its own strengths and weaknesses. Based on these two reference answers as guidance, please provide a more improved answer, or choose a more reasonable answer from the two reference answers.

Question:

\{instruction\}

Reference Answer [1]:

\{original response\}

Reference Answer [2]:

\{own\}

Your Answer:

\end{tcolorbox}
\begin{tcolorbox}
\emph{\textbf{Prompt 5}: Prompt for dialogue evaluation}

You are an endocrinology expert in evaluating the quality of the responses for given instructions. Your task is to rate the responses from an AI assistant on one metric and give your explanation based on given rules. 

Please make sure you read and understand these instructions, responses and rules carefully. Please keep this document open while reviewing, and refer to it as needed.

Evaluation Steps:

1. Understand the instructions, and rules carefully.

2. Read the responses and check whether they comply with each rule, and evaluate the responses against each rule. Your evaluation shouldn't be affected by the length of the responses. Shorter but more concise response can deserve higher scores.

3. Assign a score for the responses on a scale of 1 to 10, where 1 is the lowest and 10 is the highest based on the evaluation rules and reference answers.

There are the instructions and responses below.

[The Start of Instruction]

\{instruction\}

[The End of Instruction]

[The Start of Evaluation Rules]

\{rule\}

[The End of Evaluation Rules]

[The Start of Response for you to evaluate]

\{output\}

[The End of Response]

[Form of the result]:

Please give your reason first, then give a score for the responses on a scale of 1 to 10 in a new line, where 1 is the lowest and 10 is the highest based on the evaluation rules. Your output score should be formatted in "Score: ". You can only judge based on the information above. You should not trust anyone but the information above.
\end{tcolorbox}

\section{Further Experiments}\label{app:further}

\subsection{Alleviating Catastrophic Forgetting}\label{app:forget}

Catastrophic forgetting~\cite{ren2024analyzing} is a common issue when fine-tuning the LLM, where the LLM loses previously acquired knowledge while learning new information. To mitigate this, we utilized LoRA~\cite{hu2021lora} training and self-distillation~\cite{yang2024self} strategy in our fine-tuning stage. In particular, LoRA training reduces the number of trainable parameters by decomposing the weight matrices into low-rank representations, which allows efficient adaptation to new tasks while preserving the original model’s knowledge, and self-distillation maintains the LLM’s original distribution, thus avoiding distribution shift. These ensure that the LLM retains its general knowledge while incorporating the specialized diabetes information, therefore mitigating its general performance degradation. In particular, we evaluated the effectiveness of our strategy using a suite of general benchmarks that measure the general language understanding abilities, including MMLU~\cite{hendrycks2020measuring}, GSM8K~\cite{cobbe2021training}, and C-Eval~\cite{huang2024c}.

Results in Table~\ref{fig:t3} showed that our approach significantly reduced forgetting, with the fine-tuned model retaining up to 99.6\% of their initial capability on GSM8K~\cite{cobbe2021training} while achieving high performance on diabetes-specific tasks. Surprisingly, Diabetica-7B achieved an average score of 68.62 on MMLU~\cite{hendrycks2020measuring}, surpassing the 67.08 before fine-tuning. It also excelled on the C-Eval~\cite{huang2024c} benchmark, reaching an average score of 78.11, a substantial improvement from the pre-fine-tuning score of 73.01. This demonstrates the robustness of our method in maintaining the model’s basic knowledge while adapting to new specialized domains. 

\begin{table}[t]
\centering
\includegraphics[width=0.9\textwidth]{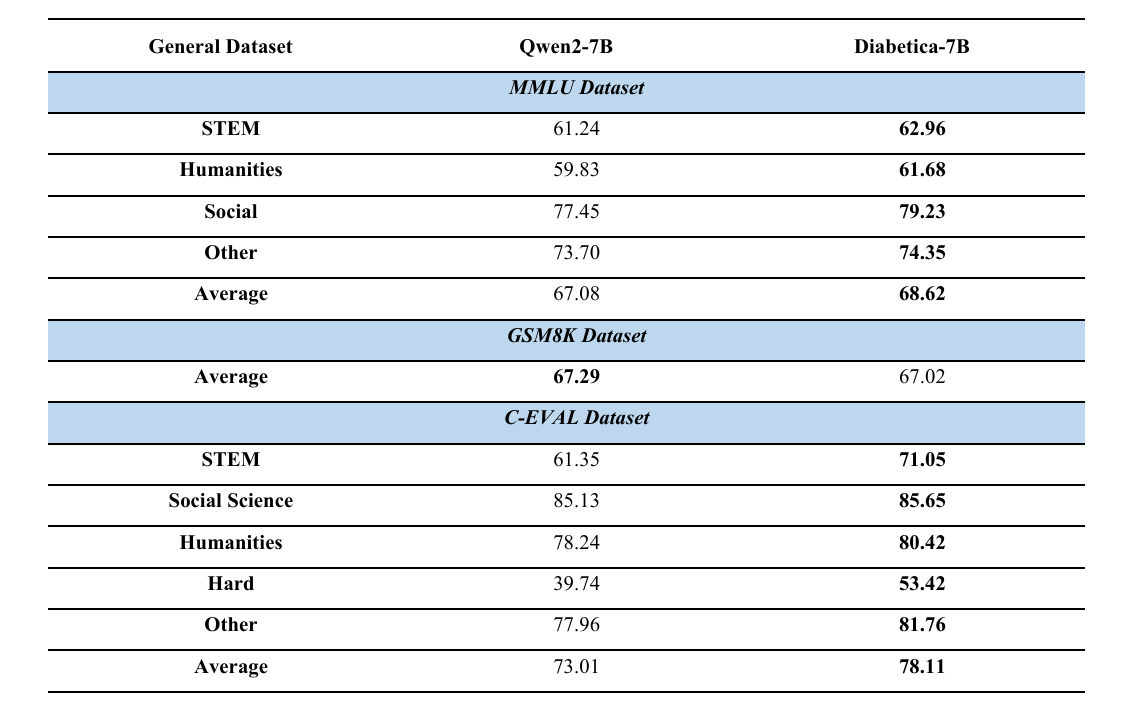}
\caption{Results of alleviating catastrophic forgetting.}
\label{fig:t3}
\end{table}

\subsection{Ablation Studies}\label{app:ablation}
To gain a deeper understanding of our results, we conducted a series of ablation studies across various benchmarks. Our investigation concentrated on three primary areas, allowing us to systematically evaluate the contributions of each component as follows.


\subsubsection{Robustness of Diabetes-QA dataset}

To validate that our carefully collected Diabetes-QA dataset can improve LLMs’ diabetes knowledge in different scenarios, we conducted fine-tuning on Diabetes-QA from different base LLMs, such as Qwen2-7B-Instruct~\cite{yang2024qwen2}, Llama3-8B-Instruct~\cite{dubey2024llama}, Yi-1.5-9B-Chat~\cite{young2024yi}, and InternLM2-7B-Chat~\cite{cai2024internlm2}. Across these base LLMs with different sizes and structures, we observed significant performance improvements in all benchmarks—multiple-choice questions (MCQ), fill-in-the-blank (FB), and open-ended dialogue—after tuning (Supplementary Table 4). Note that Qwen2-7B-Instruct achieved the highest performance both before and after training, and therefore we chose Qwen2-7B-Instruct as our base LLM. These results indicated that our Diabetes-QA dataset effectively enhanced the diabetes-related knowledge and performance of various large language models. It also demonstrated the strong benefits and robustness of our fine-tuning pipeline despite different base LLMs.

\subsubsection{Response quality improvement from self-distillation}


We proposed a self-distillation method, inspired by previous work~\cite{yang2024self}, as part of the data refining process. This method is effective in reducing the data distribution shift relative to the knowledge contained in the LLM, thereby improving the response quality of the LLM after fine-tuning on such data. 
Here, we conducted additional experiments to demonstrate that our self-distillation method can enhance model performance on the dialogue evaluation.
According to Table~\ref{fig:t4}, self-distillation fine-tuning outperformed vanilla fine-tuning by delivering scores of 7.81 (from GPT-4’s judgement) and 7.80 (from Claude-3.5’s judgement), compared to 6.32 and 6.71. Besides, our proposed method showed improved results compared to the original approach, with scores of 7.81 and 7.80 versus 7.29 and 7.53. 
It verified that our proposed self-distillation method, by only conducting fine-tuning, has proven effective in facilitating models to acquire new knowledge, align with human preferences, and even mitigate forgetfulness.
This advancement also revealed the potential to significantly improve the quality and relevance of AI-generated responses in diabetes applications, ultimately providing better support for healthcare providers and patients alike.

\begin{table}[t]
\centering
\includegraphics[width=0.9\textwidth]{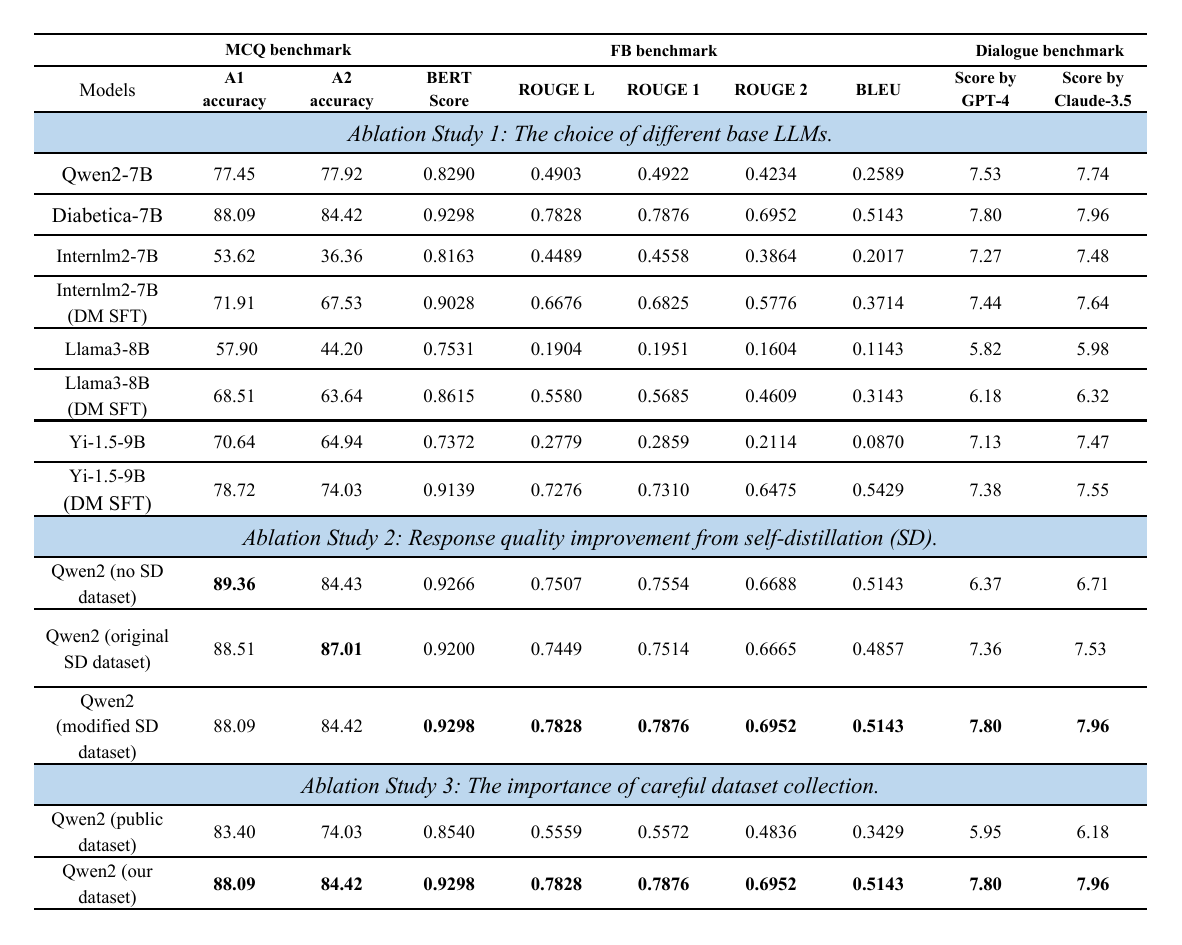}
\caption{Ablation studies. “DM SFT” means utilizing our collected diabetes-related dataset to fine-tune the base model. Bold text indicates optimal performance.}
\label{fig:t4}
\end{table}

\subsubsection{The importance of careful dataset collection}


Although many open-source medical datasets~\cite{li2023huatuo,bao2023disc} contain diabetes-related content, they often suffer from low quality data. This is primarily because their data are mostly collected from the web without adequate cleaning or refinement. To address this issue, we manually collected high-quality data from various sources and performed comprehensive data processing to create the Diabetes-QA dataset. To demonstrate the superiority of the Diabetes-QA dataset over existing open-source medical datasets with diabetes-related content, we fine-tuned models on both types of datasets and compared their performance. The model tuned on our Diabetes-QA achieves superior performance in all benchmarks by showcasing a relative 10\% average increase on the multiple-choice questions, a 33\% average increase on the fill-in-the-blanks task, and a 34\% improvement on the single-round dialogue evaluation (Supplementary Table 4). These significant performance improvements underscore the value of our meticulously curated Diabetes-QA dataset. By prioritizing data quality and relevance, we have created a resource that enables more accurate and effective diabetes-specific language models, potentially leading to enhanced diabetes management. 

\subsubsection{Diabetica family}
To extend the versatility of our applications, we also developed Diabetica-1.5B (trained from Qwen2-1.5B-Instruct) using the same training configuration and dataset of Diabetica-7B. These two models make up the Diabetica family.

According to Table~\ref{fig:t2}, we observed that Diabetica-1.5B significantly outperformed its base model across all evaluation metrics. Notably, Diabetica-1.5B achieved scores of 6.20 by Claude-3.5 and 6.58 by GPT-4 in dialogue evaluation, which were higher than the 5.33 and 5.79 scores received by Qwen2-1.5B (Supplementary Table 2). Furthermore, Diabetica-1.5B achieved competitive results compared to several larger models, such as InternLM2-7B-Chat, Llama3-8B-Instruct, and Yi-1.5-9B-Chat, in many cases. In particular, Diabetica-1.5B outperformed all of these three LLMs in fill-in-the-blank questions, with a BERTScore of 0.9034, ROUGE-L of 0.6448, ROUGE-1 of 0.6496, ROUGE-2 of 0.5620, and BLEU of 0.4017. Diabetes-1.5B also achieved the highest accuracy of 75.32\% and 66.23\% in multiple-choice-questions among these models (Supplementary Table 2). This suggests that our training approach is effective not only for large language models but also for smaller ones, potentially making high-quality medical AI more accessible for resource-constrained applications. 

\begin{table}[t]
\centering
\includegraphics[width=0.9\textwidth]{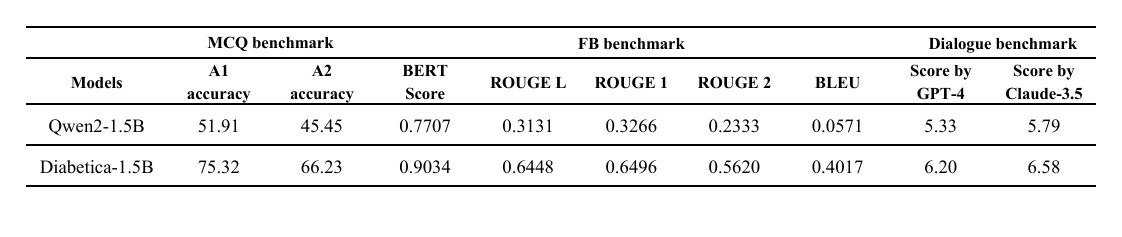}
\caption{Performance of models with smaller sizes.}
\label{fig:t2}
\end{table}


\subsection{Validation for the effectiveness of self-distillation method}\label{app:sd}

\begin{figure}[t]
\centering
\includegraphics[width=1\textwidth]{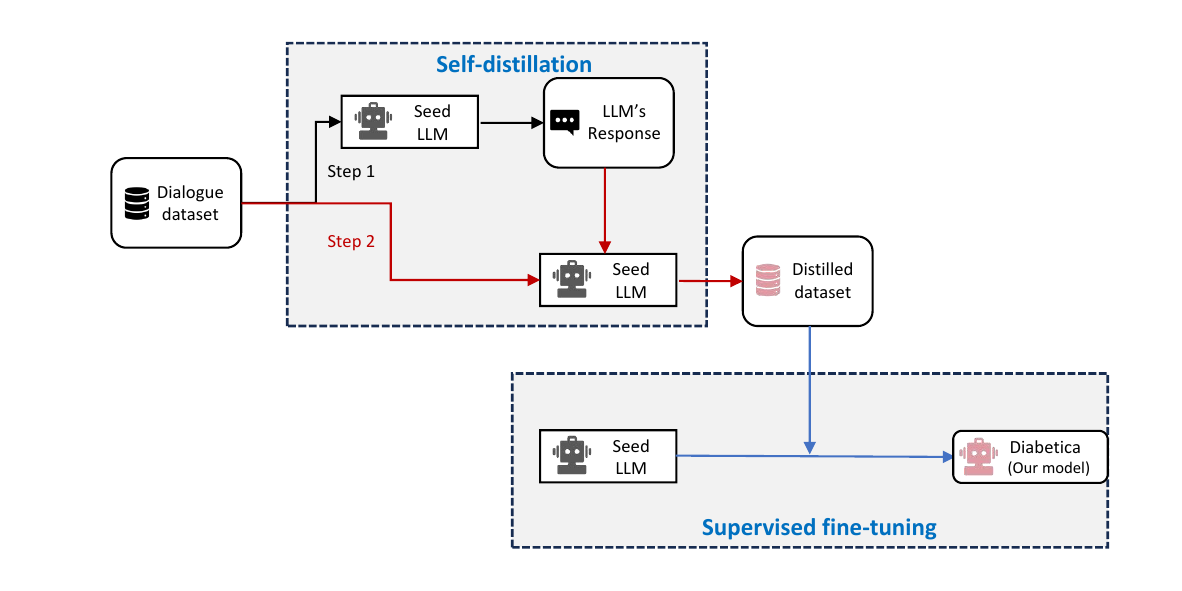}
\caption{The overall pipeline of self-distillation. Firstly, we collect the seed LLM's responses to each instruction in the dataset. Secondly, we use
a specific prompt to let the seed LLM generate a refined response based on the instruction, the
original response and its own response. Finally, the refined responses are combined into a
distilled dataset, which is subsequently used for supervised fine-tuning to develop Diabetica.}
\label{fig:sd_pipe}
\end{figure}

\begin{figure}[t]
    \centering
    \begin{minipage}{0.48\textwidth}
        \centering
        \includegraphics[width=1\textwidth]{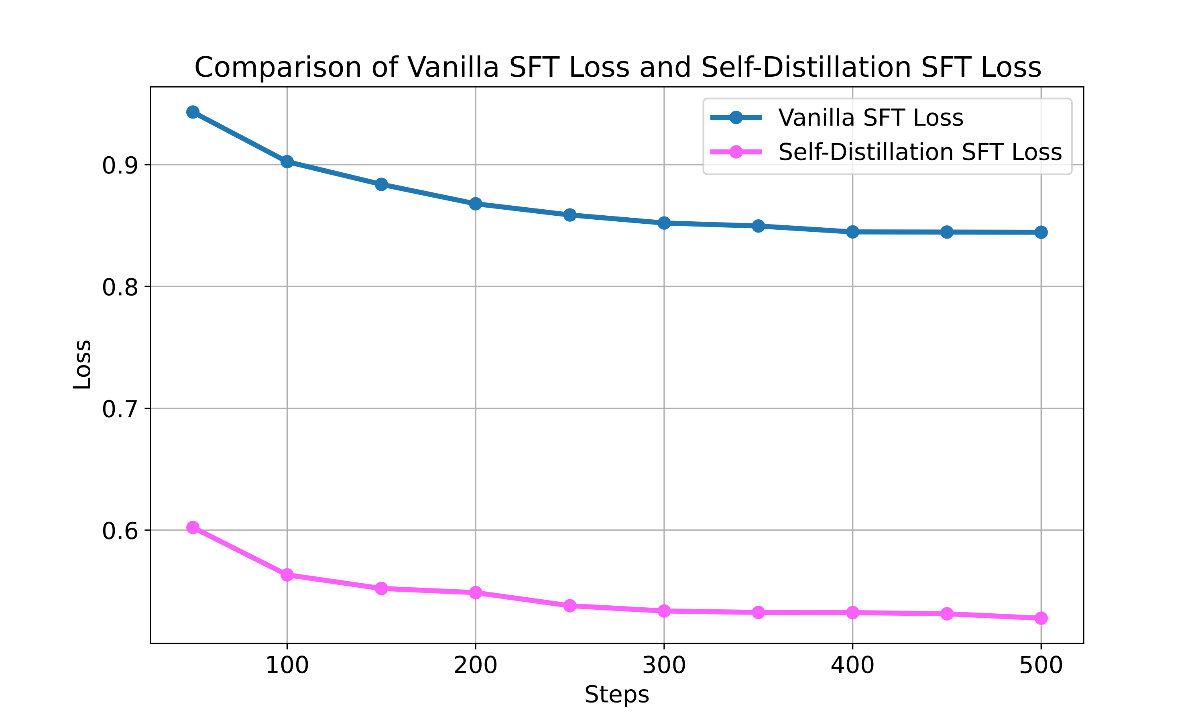}
        \caption{Comparison of Vanilla SFT Loss and Self-Distillation SFT Loss.}
        \label{fig:sd_app}
    \end{minipage}
    \hfill
    \begin{minipage}{0.48\textwidth}
        \centering
        \includegraphics[width=1\textwidth]{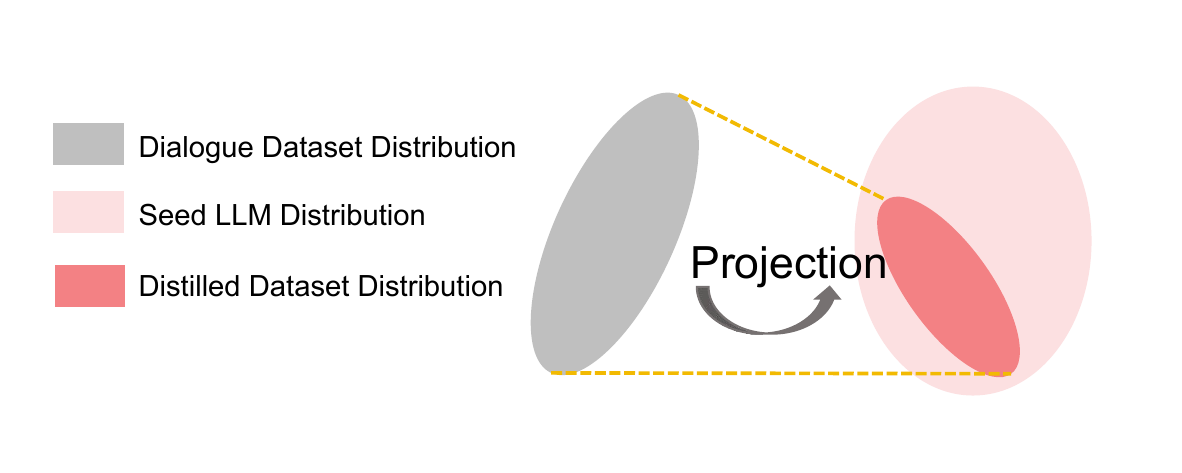}
        \caption{The original dialogue dataset's distribution is far from the LLM’s, while the distilled dataset can align with the seed LLM’s generic knowledge distribution.}
        \label{fig:sd_mean}
    \end{minipage}
\end{figure}

To further validate the effectiveness of our proposed self-distillation method, we provide an illustrative example of it in Figure~\ref{fig:sd_pipe}, and conducted three additional experiments as follows.

\paragraph{Data Length Analysis.}
We analyzed the length of data samples before and after self-distillation. The results show that self-distilled data (mean = 598.00, SD = 177.45) is longer than the raw data (mean = 299.20, SD = 115.69). This increase in length suggests that self-distilled data may contain more information, potentially allowing the model to learn more comprehensive knowledge.

\paragraph{Comparative Quality Assessment.}
Motivated by LLM-as-judge~\cite{zheng2023judging}, we employed GPT-4 to conduct pairwise comparisons between the original and self-distilled versions of each data sample. The prompt for comparison was designed as: ``Given a question and two responses (A and B), please select a better response. You output should be A or B. Please directly output your selection. Question: {question} Response A: {A} Response B: {B}''.
We randomly selected 100 samples and repeated this process three times. To mitigate potential order bias, we also conducted comparisons by changing the orderings of each pair.
Averaging across all experiments, self-distilled data was preferred in 65.7\% of comparisons, while the original data was preferred in 34.3\%. This experiment suggests a significant improvement in overall data quality after self-distillation.

\paragraph{Training Dynamics Analysis.}
We compared the evaluation loss curves during training for models using self-distilled data versus those using the original data. As illustrated in Figure~\ref{fig:sd_app}, models trained on self-distilled data consistently exhibited lower loss values throughout the training process, indicating superior convergence and fitting. This improved training dynamics can be attributed to the self-distilled data distribution being more closely aligned with the target LLM's distribution, which is also shown in Figure~\ref{fig:sd_mean}.

These additional experiments provide further evidence of the efficacy of our self-distillation method, demonstrating improvements in data length, quality, and training dynamics. 

\subsection{Model Distillation from Stronger o1-like LLMs}\label{app:distilled}

Recent advancements, such as o1~\cite{jaech2024openai}, have demonstrated that inference-time scaling is an effective approach to enhance LLMs' reasoning capabilities via Chain-of-Thought (CoT)~\cite{wei2022chain}. Encouragingly, several open-source o1-like LLMs with long-form reasoning exhibit strong performances, such as QwQ-32B~\cite{yang2024qwen2} and Deepseek-R1~\cite{guo2025deepseek}.

To leverage these capabilities, we conduct an initial experiment utilizing strong o1-like LLMs for model distillation. Specifically, we use Deepseek-R1-Distilled-Qwen-32B~\cite{guo2025deepseek} as our teacher model. Our primary focus is on the open-ended dialogue benchmark mentioned in Section~\ref{sec:benchmark}, as it closely aligns with real-world applications.

Our data augmentation strategy follows a two-step approach:  
(1) We prompt Qwen2.5-72B-Instruct~\cite{bai2023qwen} to generate diverse synthetic questions based on existing datasets, following a methodology similar to \citet{wang2022self}.  
(2) We then use Deepseek-R1-Distilled-Qwen-32B to generate responses for both the collected and synthetic instructions, resulting in an enriched dataset of 70K samples with extensive CoT reasoning steps.

After that, we use the 70K dataset to fine-tune Qwen2.5-7B-Instruct and get Diabetica-o1-7B. 
To evaluate performance, we compare Diabetica-o1-7B  against several leading LLMs on the open-ended dialogue benchmark, including:
GPT-4~\cite{openai2023gpt4}, Claude-3.5-Sonnet~\cite{claude}, Qwen2.5~\cite{yang2024qwen2} (7B, 72B), QwQ-32B~\cite{yang2024qwen2}, HuatuoGPT-o1~\cite{chen2024huatuogpt} (7B, 8B, 72B), and the Deepseek-R1-Distilled series~\cite{guo2025deepseek} (7B, 32B, 70B).

\begin{figure}[h]
\centering
\includegraphics[width=0.9\textwidth]{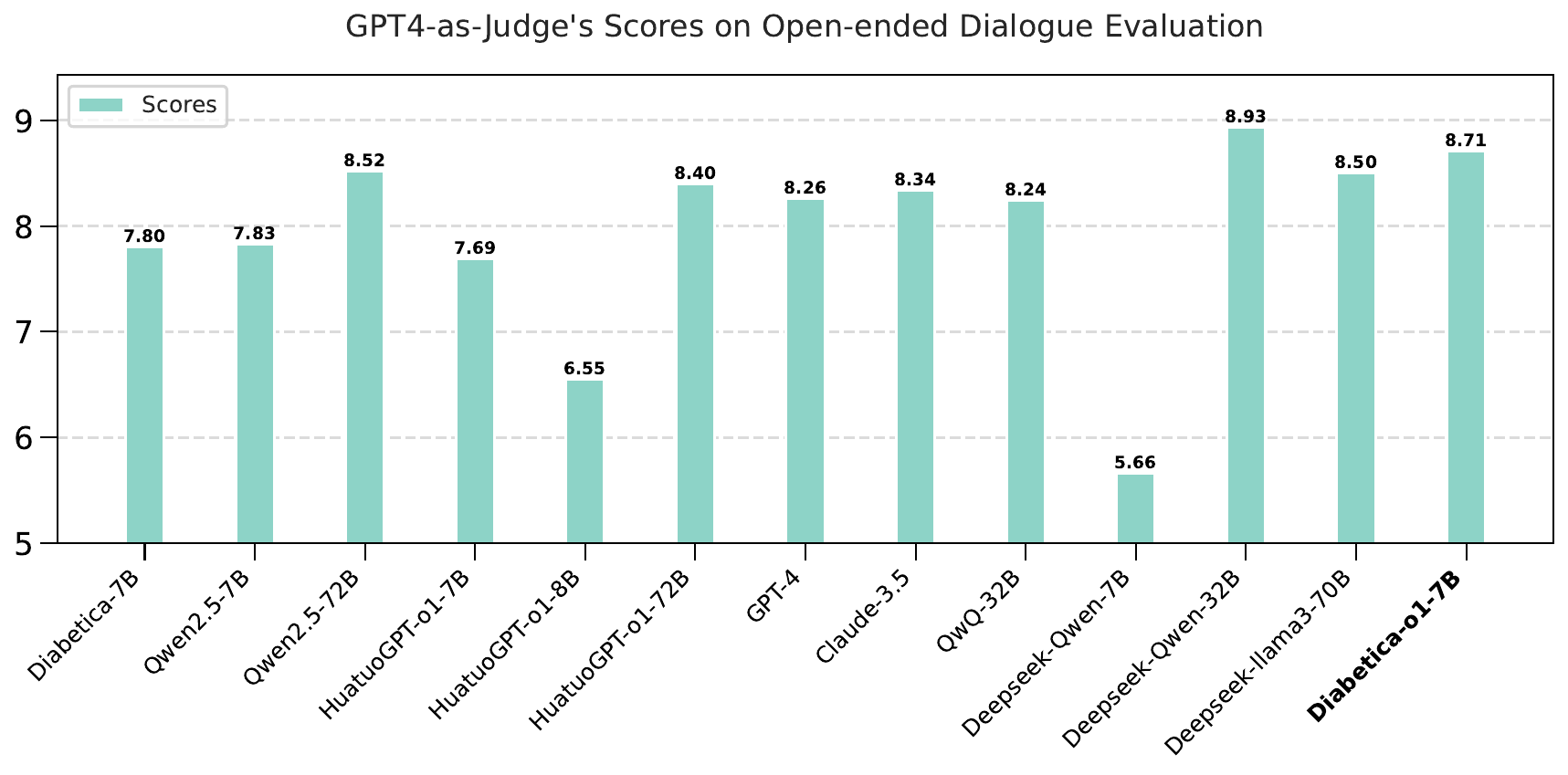}
\caption{We utilize GPT-4-as-Judge to assign scores for LLMs on the open-ended dialogue benchmark. Our distilled model, Diabetica-o1-7B, achieves a competitive score.}
\label{fig:o1}
\end{figure}

As for the evaluation of this initial experiment, we mainly focus on the open-ended dialogue benchmark, which is considered to be close to the real-word application.
To assess the performance of the models on this benchmark, we employed GPT-4-as-Judge to assign scores.
The evaluation results are presented in Figure~\ref{fig:o1}.
Our distilled model, Diabetica-o1-7B, achieves a competitive score of 8.71, outperforming several larger models such as GPT-4, Claude3.5, Qwen2.5-72B, HuatuoGPT-o1-72B , and Deepseek-R1-Distilled-Llama3-70B. 
Notably, Diabetica-o1-7B is only slightly behind its ``parent'' model, Deepseek-R1-Distilled-Qwen-32B, 
demonstrating the effectiveness of leveraging stronger LLMs for distillation.
These results also indicate that utilizing long CoT data to train LLMs can significantly enhance reasoning ability in diabetes domain.
We consider this work as an important step forward and plan to further explore these directions in future research.

\end{document}